\documentclass[pdflatex,sn-mathphys-num]{sn-jnl}

\usepackage{graphicx}%
\usepackage{multirow}%
\usepackage{amsmath,amssymb,amsfonts}%
\usepackage{amsthm}%
\usepackage[title]{appendix}%
\usepackage{xcolor}%
\usepackage{textcomp}%
\usepackage{manyfoot}%
\usepackage{booktabs}%
\usepackage{bookmark}
\usepackage{algorithm}%
\usepackage{algorithmicx}%
\usepackage{algpseudocode}%
\usepackage{listings}%
\usepackage{pdflscape}
\newcommand{\bX}{\boldsymbol{X}}
\newcommand{\bx}{\boldsymbol{x}}
\newcommand{\bbeta}{\boldsymbol{\beta}}

\theoremstyle{thmstyleone}%
\newtheorem{theorem}{Theorem}
\theoremstyle{thmstyletwo}%

\theoremstyle{thmstylethree}%

\begin{document}

\title[Article Title]{Communication-efficient distributed hazard difference estimation for heterogeneous multi-site survival data}

\author*[1,2]{\fnm{Ziwen} \sur{Wang}}
\email{wangziwen@tyut.edu.cn}
\equalcont{These authors contributed equally to this work.}
\author[2,3]{\fnm{Siqi} \sur{Li}}
\equalcont{These authors contributed equally to this work.}
\author[4]{\fnm{Marcus Eng Hock} \sur{Ong}}
\author[2,3]{\fnm{Nan} \sur{Liu}}

\affil*[1]{College of Mathematics, Taiyuan University of Technology, Taiyuan, Shanxi, China}
\affil[2]{Centre for Biomedical Data Science, Duke-NUS Medical School, Singapore}
\affil[3]{Duke-NUS AI + Medical Sciences Initiative, Duke-NUS Medical School, Singapore}
\affil[4]{Programme in Health Services and Systems Research, Duke-NUS Medical School, Singapore}

\abstract{
Multi-site collaboration can power survival models that no single hospital could fit alone, but privacy rules and protected computing environments block patient-level data sharing and the persistent server connections required by iterative federated methods.
We present DiSAH (\underline{\textbf{Di}}stributed \underline{\textbf{S}}urvival via \underline{\textbf{A}}dditive \underline{\textbf{H}}azards), a federated algorithm for time-to-event analysis whose closed-form, non-iterative structure removes the need for a dedicated central server. 
Coordination requires only aggregation of summary statistics, which any site can perform, with no patient-level data leaving the site.
DiSAH is the first federated method to estimate hazard differences, the absolute change in event rate attributable to each risk factor, providing an actionable scale for triage, resource allocation, and health-economic evaluation.
Across simulations and 47,778 emergency-department patients from the United States and Singapore, DiSAH matches centralized analysis in accuracy and discrimination, recovers mortality risk factors no individual site was powered to detect, and outperforms meta-analysis and local models.
}

\keywords{Distributed additive hazards model; Hazard difference effects; Absolute risk; Multi-site survival data; Federated learning; Privacy-preserving estimation}

\maketitle

\section{Introduction}\label{sec1}

The widespread adoption of electronic health records (EHRs) has made large-scale prognostic modeling possible, yet the data to support it remain locked inside individual institutions~\citep{Greenhalgh2008, Shea2010}. Pooling patient records across hospitals increases statistical power, improves generalizability, and makes rare outcomes estimable, advantages that matter most for time-to-event analyses, where events are scarce at any single site. Two barriers stand in the way. Privacy regulations and institutional policies prohibit moving patient-level data across sites~\citep{Maro2009}, and the protected computing environments that hold clinical data increasingly sit behind firewalls that forbid the persistent external connections many federated methods depend on~\citep{li2024federated}. A practical method must therefore respect both constraints at once, learning across sites without centralizing data and without requiring a standing connection to a central server.

Several distributed methods have been developed for survival analysis under the Cox proportional hazards framework. Model-agnostic federated learning approaches such as FedAvg~\citep{mcmahan2017communication} and FedProx~\citep{li2020federated} require persistent server connections and iterative optimization rounds, which are impractical in clinical environments where hospital data reside behind institutional firewalls that prohibit external connections~\citep{li2024federated}.
These methods are also designed for prediction rather than valid inference on covariate effects. Model-specific alternatives, including ODAC~\citep{Duan2020} and ODACH~\citep{luo2022odach, LiD2023}, remove the need for persistent connections through Taylor expansion--based approximations that require only summary-level statistics. All of these methods, however, operate within the Cox model's multiplicative hazard framework, yielding hazard \emph{ratios} (relative risk measures).

In many clinical and public health contexts, hazard ratios alone are not always sufficient for decision-making. The \emph{hazard difference}, defined as the absolute change in the instantaneous event rate per unit change in a covariate~\citep{lin1994, Martinussen2002AH}, directly quantifies how much a risk factor increases or decreases the event rate on an absolute scale. Absolute risk measures are often preferred for clinical interpretation and health-economic evaluation because they convey the magnitude of risk in natural units, enabling direct comparison of the public health impact across exposures and populations~\citep{Zhang2018RD}. Estimation of hazard differences under additive hazards models from single-site data is well established~\citep{lin1994, Yin2004AH, Wang2010AH, Zhao2025AH}, but no distributed or federated method currently exists for this estimand. Table~\ref{tab:method-comparison} summarizes this gap relative to existing approaches.

\begin{table*}[!htbp]
\centering
\footnotesize
\setlength{\tabcolsep}{4pt}
\caption{Comparison of federated survival analysis methods. $\checkmark$ = supported, $\times$ = not supported.}
\begin{tabular}{lccccc}
\toprule
Property & ODAC & ODACH  & FedAvg$^*$ &\textbf{DiSAH (Ours)} \\
 & \citep{luo2022odach} & \citep{LiD2023} & \citep{mcmahan2017communication} & \\
\midrule
Hazard Differences (Absolute Risk) & $\times$ & $\times$ & $\times$ & $\checkmark$ \\
Hazard Ratio (Relative Risk) & $\checkmark$ & $\checkmark$ & $\times$ & $\times$ \\
Valid Statistical Inference & $\checkmark$ & $\checkmark$ & $\times$ & $\checkmark$ \\
Server-independent & $\checkmark$ & $\checkmark$ & $\times$ & $\checkmark$\\
Communication Rounds & 1 & 1 & iterative & 1 or 3 \\
\bottomrule
\end{tabular}

\vspace{1mm}
{\scriptsize 
$^*$FedAvg represents engineering-based FL methods broadly (e.g., FedProx~\citep{li2020federated}, FedAvgM~\citep{hsu2019measuring}); listed as a representative example.
}
\label{tab:method-comparison}
\end{table*}

In this work, we propose DiSAH (\underline{\textbf{Di}}stributed \underline{\textbf{S}}urvival via \underline{\textbf{A}}dditive \underline{\textbf{H}}azards), a communication-efficient, privacy-preserving algorithm for estimating hazard differences from multi-site survival data. DiSAH comprises two estimators. An unstratified variant (DiSAH-U) for settings where sites share a common baseline hazard, requiring three rounds of summary-level statistics, and a stratified variant (DiSAH-S) that accommodates site-specific baseline hazards in a single communication round. Neither variant requires patient-level data sharing, persistent server connections, or iterative optimization. We establish asymptotic normality for both estimators and prove that DiSAH-U is asymptotically equivalent to the pooled oracle estimator. 

We evaluate DiSAH through simulations and a real-world application using heterogeneous EHR data from emergency departments in the United States (MIMIC-IV-ED) and Singapore (Singapore General Hospital). DiSAH achieves estimation accuracy comparable to pooled analysis while consistently outperforming meta-analysis and local site-specific models, and identifies clinically significant risk factors for 30-day mortality that individual-site analyses lacked power to detect.

\section{Methods}\label{sec11}
\subsection{Preliminaries}
Let non-negative random variable $T$ denote the event time of interest and let $\bX \in \mathbb{R}^p$ be a vector of covariates.
The \emph{additive hazards model} specifies the hazard at time $t$ as:
\begin{equation}\label{m1}
  \lambda(t \mid \bX)=\lambda_{0}(t)+\bbeta^{T}\bX,
\end{equation}
where $\lambda_{0}(t)$ is an unspecified baseline hazard function, and $\bbeta \in \mathbb{R}^p$ is the coefficient vector representing the \emph{hazard differences} (HD), i.e., the absolute change in hazard per unit change in each covariate. 
This contrasts with the Cox model's multiplicative structure $\lambda(t \mid \bX) = \lambda_0(t)\exp(\bbeta^\top \bX)$, which yields hazard \emph{ratios}.

For right-censored survival data on individuals $i = 1, \ldots, n$, we observe $(y_i, \delta_i, \bx_i)$,  where $y_i = \min(T_i, C_i)$ is the observed time, $C_i$ is the censoring time and $\delta_i = \mathbf{1}(T_i \leq C_i)$ indicates whether the event was observed. 
Let $N_i(t) = \mathbf{1}(y_i \leq t, \delta_i = 1)$ denote the counting process and $Y_{i}(t)=I(y_i \geq t)$ the at-risk process, where $Y_{i}(t)$ takes value 1 indicating that subject $i$ is within the risk set $R(t)=\{i: y_i \geq t\}$.
Under standard independence assumptions that $C$ is independent of $T$ and $\bX$, and that $T$ and $C$ are conditionally independent given $\bX$, $\bbeta$ can be estimated via the martingale-based estimating equation~\citep{lin1994} for the model in Eq.~\eqref{m1}.

Specifically, the score function is given by
\begin{equation}\label{m2}
  U(\bbeta) =
  \sum_{i=1}^{n}\int_{0}^{\tau}\big\{\bx_{i}-\bar{\bx}(t)\big\}\{dN_{i}(t)-\bbeta^{T}\bx_{i}Y_{i}(t)dt \},
\end{equation}
where $\bar{\bx}(t)=\sum_{i=1}^{n}\bx_{i} Y_i(t)~/~ \sum_{i=1}^{n}Y_i(t)$ represents the average covariates vector over the risk set at time $t$, and $\tau$ is a prespecified sufficiently large constant satisfying $P(Y\geq \tau)>0$. In practice, $\tau$ is often taken as the latest observed time in the study.

Solving $U(\bbeta)=0$ yields a closed-form estimator $A^{*-1}D^{*}$, where $A^{*}=n^{-1}\sum_{i=1}^{n}\int_{0}^{\tau}Y_{i}(t)\big\{\bx_{i}-\bar{\bx}(t)\big\}^{\otimes 2}dt$
\footnote{Throughout this paper, we use the notation $a^{\otimes 2}$ to denote the outer product $aa^{T}$ for a vector $a$.}, 
and
$D^{*}=n^{-1}\sum_{i=1}^{n}\int_{0}^{\tau}\big\{\bx_{i}-\bar{\bx}(t)\big\}dN_{i}(t)$.
The covariance matrix can be estimated by $A^{*-1}\Sigma^{*}A^{*-1}$, where $\Sigma^{*}=n^{-1}\sum_{i=1}^{n}\int_{0}^{\tau}\big\{\bx_{i}-\bar{\bx}(t)\big\}^{\otimes 2}dN_{i}(t)$.
Since $dN_i(t)$ equals zero except at the failure time of the $i$th subject, $D^{*}$ simplifies to $\tilde{D}=n^{-1}\sum_{i=1}^{n}\delta_{i}\{\bx_{i}-\bar{\bx}(y_{i})\}$.
Similarly we rewrite $\Sigma^{*}$ as $\tilde{\Sigma}=n^{-1}\sum_{i=1}^{n}\delta_{i}\{\bx_{i}-\bar{\bx}(y_{i})\}^{\otimes 2}$.

In practice, $A^{*}$ can be approximated by a Riemann sum. Let observation times be ordered as $y_{(1)}\leq y_{(2)}\leq\cdots\leq y_{(n)}$
and denote $\Delta y_{(i)}=y_{(i)}-y_{(i-1)}$, where $y_{(0)}=0$.
Then $A^{*}$ can be approximated by
$\tilde{A}=n^{-1}\sum_{i=1}^{n}\sum_{l\in R(y_{(i)})}\big\{\bx_{l}-\bar{\bx}(y_{(i)})\big\}^{\otimes 2}\Delta y_{(i)}$.
The HD coefficient vector can then be estimated by
\begin{equation}\label{m3}
  \hat{\beta}_{G}=\tilde{A}^{-1}\tilde{D}.
\end{equation}
with covariance matrix
\begin{equation}\label{mm3}
 \hat{I}_{G}= \tilde{A}^{-1}\tilde{\Sigma}\tilde{A}^{-1}.
\end{equation}

\begin{figure*}[t]
  \begin{center}
    \centerline{\includegraphics[width=1\textwidth]{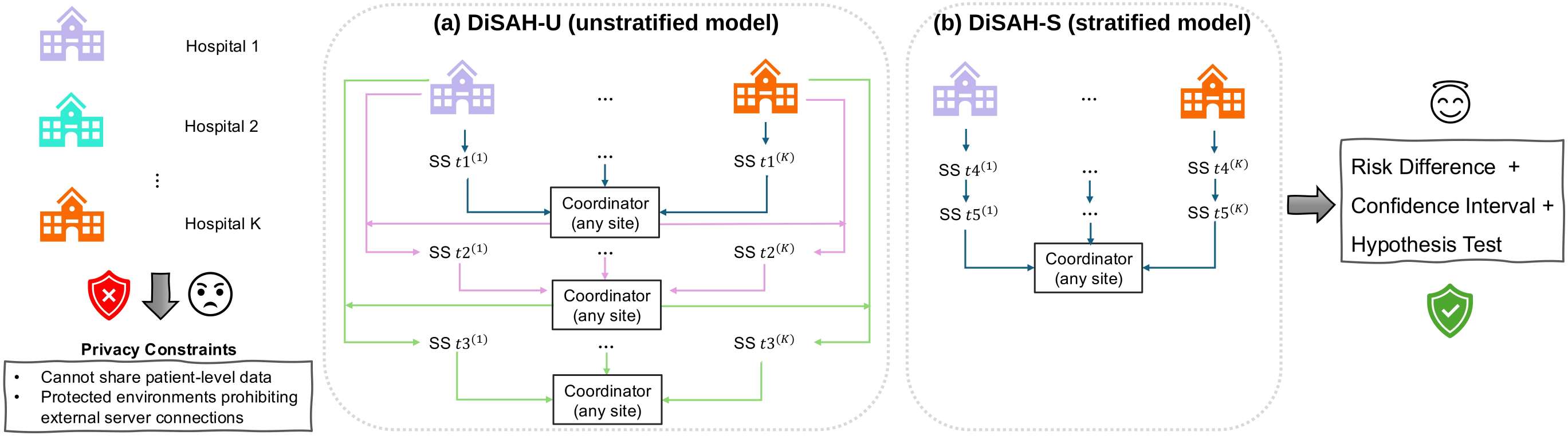}}
    \caption{Workflow of DiSAH. Each site computes and shares only summary-level statistics (SS); the exact forms are described in the Methods section. The coordinator performs only aggregation and broadcasting without iterative optimization and can be any participating site; no central server is required.}
    \label{workflow}
  \end{center}
\end{figure*}

\subsection{Distributed hazard difference estimation via unstratified additive hazards model}
\label{subsec.fedrd.unstratified}

We first consider the homogeneous setting under the unstratified additive hazards model, assuming that all sites share a common baseline hazard $\lambda_{0}(t)$, i.e., each site satisfies the model in Eq.~\eqref{m1}.

Consider a multi-site survival study involving $K$ independent data-contributing sites (e.g., hospitals). 
Suppose the total sample comprises $n$ independent individuals across the $K$ sites. 
For each site $k=1,...,K$, let $\mathcal{O}_k=\{i: i\ \text{in site}\ k, \text{for}\ i=1,...,n \}$ be the index set for the $k$-th site with sample size $n_k$  so that $n=n_1+n_2+\cdots +n_k$.
For the $i$-th patient at the $k$-th site, we observe $\{(y_{ik}; \delta_{ik}; \bx_{ik}), k=1,...,K; i=1,...,n_{k}\}$. 
If patient-level data could be pooled across all $K$ sites, the pooled estimator in Eq.~\eqref{m3} would serve as the ground truth; however, privacy constraints prevent direct data sharing, motivating our federated approach.

\begin{algorithm}[!t]
\caption{DiSAH-U (unstratified, 3 communication rounds)}\label{A1}
\begin{algorithmic}[1]
\Statex \textbf{At each site} $k = 1, \dots, K$
\State Sort $\{y_{lk};\; l \in \mathcal{O}_{k}\}$ to $\{y_{(l)k};\; l \in \mathcal{O}_{k}\}$
\State Broadcast $\{y_{(l)k};\; l \in \mathcal{O}_{k}\}$ to other sites
\Statex
\Statex \textbf{At the coordinator$^\dagger$}
\State Integrate and sort $\{y_{(l)1}\}, \dots, \{y_{(l)K}\}$ to form $\{y_{(i)},\; i \in \mathcal{O}\}$
\State Send $\{y_{(i)},\; i \in \mathcal{O}\}$ to each site
\Statex
\Statex \textbf{At each site} $k = 1, \dots, K$
\State Calculate item (t2): $\sum_{l=1}^{n_{k}}\bx_{lk}I(y_{lk}\geq y_{(i)})$ and $\sum_{l=1}^{n_{k}}I(y_{lk}\geq y_{(i)})$
\State Transfer item (t2) to the coordinator
\Statex
\Statex \textbf{At the coordinator}
\State Obtain $\bar{\bx}(y_{(i)})$ by integrating item (t2)
\State Send $\bar{\bx}(y_{(i)})$ and $\{y_{(i)},\; i \in \mathcal{O}\}$ to each site
\Statex
\Statex \textbf{At each site} $k = 1, \dots, K$
\State Calculate item (t3) based on $\bar{\bx}(y_{(i)})$ and $\{y_{(i)},\; i \in \mathcal{O}\}$
\State Transfer item (t3) to the coordinator
\Statex
\Statex \textbf{At the coordinator}
\State Obtain the estimator $\hat{\bbeta}_{U}$ based on the closed form~\eqref{m5}
\Statex
\Statex {\footnotesize $\dagger$The coordinator requires only aggregation and broadcasting. Any participating site can serve as coordinator; no central server is required.}
\end{algorithmic}
\end{algorithm}

In the multi-site settings, define $R_{k}(t) = \{i: y_{ik} \geq t \}$ as the risk set at time $t$ from site $k$. The combined risk set over all the $K$ sites is $R(t)= \{(i,k); y_{ik}\geq t\}$, and the counting process $N_{ik}(t)=I(y_{ik}\leq t, \delta_{ik}=1)$. The score function (Eq.~\eqref{m2}) can then be rewritten as
\begin{equation}\label{m4}
U(\bbeta)=\sum_{k=1}^{K}\sum_{i=1}^{n_{k}}\int_{0}^{\tau}\big\{\bx_{ik}-\bar{\bx}(t)\big\}\{dN_{ik}(t)-\bbeta^{T}\bx_{ik}Y_{ik}(t)dt \},
\end{equation}
where $Y_{ik}(t)$ is a $0-1$ at-risk process, taking value $1$ if subject $i$ from site $k$ is at risk at time $t$, i.e., $(i,k) \in R(t)$. 
The term $\bar{\bx}(t)$ can also be rewritten as $\sum_{i=k}^{K}\sum_{i=1}^{n_k}Y_{ik}(t)\bx_{ik} / \sum_{k=1}^{K}\sum_{i=1}^{n-k}Y_{ik}(t)$, which equals to $\sum_{(i,k) \in R(t)}\bx_{ik} / |R(t)|$.
As in the single-site case, we approximate the integration by a Riemann sum to obtain a closed-form estimator:
\begin{align}
\hat{\bbeta}_{U} &=\Big[\sum_{i=1}^{n}\sum_{(l,k) \in R(y_{(i)})} \big\{\bx_{lk}-\bar{\bx}(y_{(i)}) \big\}^{\otimes 2}\Delta y_{(i)} \Big]^{-1} \nonumber\\
 &\quad\ 
 \times
 \Big[\sum_{k=1}^{K}\sum_{i=1}^{n_{k}}\delta_{ik}\big\{\bx_{ik} -\bar{\bx}(y_{ik})  \big\} \Big] \nonumber \\
   &= \Big[\sum_{k=1}^{K}\sum_{l=1}^{n_k}\sum_{i=1}^{n}I(y_{lk}\geq y_{(i)}) \big\{\bx_{lk}-\bar{\bx}(y_{(i)}) \big\}^{\otimes 2} \Delta y_{(i)} \Big]^{-1} \nonumber\\
   &\quad\ 
   \times
   \Big[\sum_{k=1}^{K}\sum_{i=1}^{n_{k}}\delta_{ik}\big\{\bx_{ik} -\bar{\bx}(y_{ik})  \big\} \Big], \label{m5}
\end{align}
and the $\tilde{\Sigma}$ part of the covariance matrix in Eq.~\eqref{mm3} can be estimated by $n^{-1}\sum_{k=1}^{K}\sum_{i=1}^{n_{k}}\delta_{ik}\big\{\bx_{i} -\bar{\bx}(y_{i}) \big\}^{\otimes 2}$.

In sum, to construct the HD estimator $\hat{\bbeta}_{U}$ and its covariance matrix $\hat{I}_{U}$, each data-contributing site $k=1,...,K$ needs to provide the following:
 \begin{itemize}
 \item[(t1)] $\{y_{lk}; l \in \mathcal{O}_{k}\}$;
 \item[(t2)] $\sum\limits_{l=1}^{n_{k}}\bx_{lk}I(y_{lk}\geq y_{(i)})$, and $\sum\limits_{l=1}^{n_{k}}I(y_{lk}\geq y_{(i)})$;

  \item[(t3)]   $\sum\limits_{l=1}^{n_{k}}\sum\limits_{i=1}^{ n}I(y_{lk}\geq y_{(i)})\big\{ \bx_{lk}-\bar{\bx}(y_{(i)}) \big\}^{\otimes 2}\Delta y_{(i)}$, 

$\sum\limits_{l=1}^{n_{k}}\delta_{lk}\big\{ \bx_{lk}-\bar{\bx}(y_{lk}) \big\}$;

and
$\sum\limits_{l=1}^{n_{k}}\delta_{lk}\big\{ \bx_{lk}-\bar{\bx}(y_{lk}) \big\}^{\otimes 2}$.
\end{itemize}

Notably, the $k$-th site must share the observed time set, $\{y_{lk}; l \in \mathcal{O}_{k}\}$. To mitigate potential privacy leakage, each site provides only the ordered observed times without the individual labels, i.e. $\{y_{(l)k}; l \in \mathcal{O}_{k}\}$, from which the global ordered times $\{y_{(i)}, i \in \mathcal{O}\}$ are obtained.
A step-by-step implementation of DiSAH-U is described in Algorithm~\ref{A1}, with an illustration in 
Figure~\ref{workflow}(a).

\subsection{Distributed hazard difference estimation via stratified additive hazards model}
To account for inter-site heterogeneity, we consider a stratified additive hazards model and propose DiSAH-S.
In this framework, the baseline hazards differ across sites while $\bbeta$ remains common, i.e., the covariate effects on time-to-event are shared across sites.
The stratified additive hazards model for site $k$ is defined as:
\begin{equation}\label{stratified_mk}
  \lambda_{ik}(t|\bx_{ik})= \lambda_{0k}(t)+\bbeta^{T}\bx_{ik},
\end{equation}
where $\lambda_{0k}(t)$ is the baseline hazard function for the $k$-th site.

Following the same arguments used to derive \eqref{m2}, score function based on the data of the $k$-th site can be expressed as
\begin{equation}\label{m6}
  S_{k}(\bbeta)=\sum_{i=1}^{n_{k}}\int_{0}^{\tau}\big\{\bx_{ik}-\bar{\bx}_{k}(t)\big\}\{dN_{ik}(t)-\bbeta^{T}\bx_{ik}Y_{ik}(t)dt \}
\end{equation}
where $\bar{\bx}_{k}(t)=\sum_{i=1}^{n_k}Y_{ik}(t)\bx_{ik} ~/~ \sum_{i=1}^{n_k}Y_{ik}(t)$ and equals to $\sum_{i \in R_k(t)}\bx_{ik} ~/~ |R_k(t)|$ with $R_k(t)= \{i; y_{ik}\geq t\}$.

Let the observation times of site $k$ be ordered as $y_{(1)k}\leq y_{(2)k}\leq\cdots\leq y_{(n_k)k}$ and denote $\Delta y_{(i)k}=y_{(i)k}-y_{(i-1)k}$, where $y_{(0)k}=0$.
Solving the estimating equation $S_{k}(\bbeta)=0$ yields a consistent local estimator of $\bbeta$, denoted by $\hat{\bbeta}_{k}=A_{k}^{-1}D_{k}$ for $k=1,...,K$, where $A_k=n_{k}^{-1}\sum_{i=1}^{n_k}\sum_{l\in R_k(y_{(i)k})}\big\{\bx_{lk}-\bar{\bx}_k(y_{(i)k})\big\}^{\otimes 2}\Delta y_{(i)k}$ and 
$D_k= n_{k}^{-1}\sum_{i=1}^{n_k}\delta_{ik}\{\bx_{ik}-\bar{\bx}_{k}(y_{ik})\}$.
The covariance matrix of the local estimator is estimated by $A_{k}^{-1}\Sigma_{k}A_{k}^{-1}$,  where $\Sigma_{k}=n_{k}^{-1}\sum_{i=1}^{n_{k}}\delta_{ik}\big\{\bx_{i} -\bar{\bx}(y_{i}) \big\}^{\otimes 2}$.

Combining the $K$ site-specific score functions, the overall score function is
\begin{equation}\label{m_Sbbeta}
  S(\bbeta)=\sum\limits_{k=1}^{K} S_{k}(\bbeta)
\end{equation}
Letting $p_k=n_k/n$, we can approximate $\bbeta$ by
\begin{align}
 \hat{\bbeta}_{S} &=\Big\{\sum\limits_{k=1}^{K}p_{k}A_k\Big\}^{-1}\Big\{\sum\limits_{k=1}^{K}p_{k}D_k\Big\}  \nonumber \\
   &=\Big\{\sum_{k=1}^{K}\sum_{i=1}^{n_k}\sum\limits_{l \in R_k(y_{(i)k})}\big\{\bx_{lk}-\bar{\bx}_{k}(y_{(i)k})\big\}^{\otimes 2}\Delta y_{(i)k} \Big\}^{-1} \nonumber\\
   &\quad\ \Big\{\sum_{k=1}^{K}\sum_{i=1}^{n_k}\delta_{ik}\big\{\bx_{ik}-\bar{\bx}_{k}(y_{ik})\big\}\Big\}. \label{m8}
\end{align}

To obtain the estimator by setting Eq.~\eqref{m6} to zero, site-specific items (summary-level statistics) are transferred to the coordinator to obtain Eq.~\eqref{m8}:
 \begin{itemize}
 \item[(t4)] $\sum\limits_{i=1}^{n_{k}}\bx_{ik}I(y_{ik}\geq y_{(i)k})$, 

 and $\sum\limits_{i=1}^{n_{k}}I(y_{ik}\geq y_{(i)k})$.
  \item[(t5)]   $\sum\limits_{i=1}^{n_{k}}\sum\limits_{l \in R_{k}(y_{(i)k})}\big\{ \bx_{lk}-\bar{\bx}_{k}(y_{(i)k}) \big\}^{\otimes 2}\Delta y_{(i)k}$, 
  
    $\sum\limits_{i=1}^{n_{k}}\delta_{ik}\big\{ \bx_{ik}-\bar{\bx}_{k}(y_{ik}) \big\}$, 
    
    and
    $\sum\limits_{i=1}^{n_{k}}\delta_{ik}\big\{ \bx_{ik}-\bar{\bx}_{k}(y_{ik}) \big\}^{\otimes 2}$.
\end{itemize}

The implementation is summarized in Algorithm~\ref{A2}, with an illustration in Figure~\ref{workflow} (b). Unlike DiSAH-U, which requires three communication rounds, DiSAH-S completes in a single round.

\begin{algorithm}[!t]
\caption{DiSAH-S (stratified, 1 communication round)}\label{A2}
\begin{algorithmic}[1]
\Statex \textbf{At each site} $k = 1, \dots, K$
\State Calculate items (t4) and (t5), and send item (t5) to the coordinator
\Statex
\Statex \textbf{At the coordinator}
\State Obtain the estimator $\hat{\bbeta}_{S}$ based on the closed form in Eq.~\eqref{m8}
\end{algorithmic}
\end{algorithm}

\begin{table*}[ht]
  \caption{Simulation results for the hazard difference estimation when $n_1=n_2=n_3=n_4=n_5=100$ (configuration (i), balanced), with computation time in seconds.}\label{Tab.balanced}
   \setlength{\tabcolsep}{0.06cm}
   \centering
   \footnotesize
   \begin{tabular*}{\textwidth}{@{\extracolsep\fill}lcccccccccccccc@{\extracolsep\fill}}
   \toprule
   & &  \multicolumn{6}{@{}c@{}}{Scenario 1 (homogeneous)} & & \multicolumn{6}{@{}c@{}}{Scenario 2 (heterogeneous)}\\
   \cmidrule{3-8} \cmidrule{10-15}
   Methods&Para&Bias&SD&SE&CP&MSE&Runtime& &Bias &SD &SE &CP&MSE&Runtime\\
   \midrule
   Pooled   &$\beta_1$  & 0.029 & 0.317 & 0.319 & 0.948 & 0.101 & 0.054    &   & 0.028 & 0.335 & 0.336 & 0.950 & 0.113 & 0.057 \\
             &$\beta_2$   & 0.002 & 0.308 & 0.314 & 0.942 & 0.095 &             &  & 0.003 & 0.334 & 0.331 & 0.944 & 0.111 & \\
             & $\beta_3$   & 0.002 & 0.179 & 0.184 & 0.966 & 0.032 &             &  & $-$0.003 & 0.187 & 0.193 & 0.956 & 0.035 &  \\
   \addlinespace
   DiSAH-U    &$\beta_1$  & 0.029 & 0.317 & 0.319 & 0.948 & 0.101 & 1.056 & & 0.028 & 0.335 & 0.336 & 0.950 & 0.113 & 1.063 \\
                       & $\beta_2$   & 0.002 & 0.308 & 0.315 & 0.942 & 0.095 &   &    & 0.003 & 0.334 & 0.331 & 0.944 & 0.111 &   \\
                        & $\beta_3$   &0.002 & 0.179 & 0.184 & 0.966 & 0.032 &   &   & $-$0.003 & 0.187 & 0.193 & 0.956 & 0.035 &   \\
   \addlinespace
   DiSAH-S  &$\beta_1$    & 0.030 & 0.318 & 0.322 & 0.950 & 0.102 & 0.130  & & 0.024 & 0.330 & 0.338 & 0.956 & 0.109 & 0.131  \\
                      & $\beta_2$    & 0.004 & 0.311 & 0.317 & 0.958 & 0.096 &   &     & $-$0.001 & 0.332& 0.333 & 0.944 & 0.110 &   \\
                      &$\beta_3$    & 0.003 & 0.179 & 0.185 & 0.956 & 0.032    &    &   & 0.000 & 0.185 & 0.194 & 0.960 & 0.034 &    \\
   \addlinespace
  Meta    &$\beta_1$   & 0.060 & 0.334 & 0.371 & 0.958 & 0.115 & 0.042  & & 0.054 & 0.357 & 0.397 & 0.968 & 0.130 & 0.042   \\
            &  $\beta_2$   & 0.015 & 0.327 & 0.366 & 0.974 & 0.107 &     & & 0.009 & 0.360 & 0.392 & 0.970 & 0.129 &  \\
            &  $\beta_3$   & 0.015 & 0.184 & 0.212 & 0.986 & 0.034 &   & & 0.012 & 0.197 & 0.227 & 0.972 & 0.039 &  \\
   \addlinespace
 Local 1   &$\beta_1$   & 0.099 & 0.758 & 0.738 & 0.934 & 0.583 & 0.009   &     & 0.099 & 0.758 & 0.738 & 0.934 & 0.583 & 0.009  \\
   &          $\beta_2$    & $-$0.026 & 0.744 & 0.732 & 0.958 & 0.553 &     &   & $-$0.026 & 0.744 & 0.732 & 0.958 & 0.553 &  \\
   &          $\beta_3$    & 0.024 & 0.420 & 0.423 & 0.964 & 0.177 &   &   & 0.024 & 0.420 & 0.423 & 0.964 & 0.177 &    \\
   \addlinespace
 Local 2   &$\beta_1$  & $-$0.010 & 0.710 & 0.739 & 0.960 & 0.503 & 0.008    &   & $-$0.011 & 0.653 & 0.680 & 0.962 & 0.426 & 0.008 \\
   &          $\beta_2$  & $-$0.021 & 0.733 & 0.728 & 0.960 & 0.537 &    & & $-$0.025 & 0.679 & 0.668 & 0.962 & 0.461 & \\
   &          $\beta_3$  & 0.012 & 0.436 & 0.425 & 0.960 & 0.190 &   &  & 0.013 & 0.400 & 0.391 & 0.960 & 0.160 &    \\
   \addlinespace
 Local 3   &$\beta_1$  & 0.094 & 0.776 & 0.744 & 0.942 & 0.610 & 0.008   & & 0.080 & 0.686 & 0.655 & 0.948 & 0.476 & 0.008  \\
   &          $\beta_2$  & 0.066 & 0.776 & 0.736 & 0.936 & 0.605 &     && 0.056 & 0.664 & 0.647 & 0.946 & 0.443 &\\
   &          $\beta_3$  & $-$0.005 & 0.425 & 0.425 & 0.962 & 0.181 &   &   & $-$0.007 & 0.375 & 0.374 & 0.964 & 0.140 &   \\
   \addlinespace
 Local 4   &$\beta_1$  & 0.024 & 0.731 & 0.749 & 0.956 & 0.534 & 0.008 & & 0.011 & 0.955 & 0.970 & 0.956 & 0.910 & 0.008 \\
   &          $\beta_2$   & $-$0.028 & 0.706 & 0.737 & 0.966 & 0.498 &    & & $-$0.042 & 0.919 & 0.956 & 0.972 & 0.845 &  \\
   &          $\beta_3$   & 0.016 & 0.443 & 0.427 & 0.940 & 0.196 &    & & 0.002 & 0.577 & 0.551 & 0.946 & 0.332 &    \\
   \addlinespace
   Local 5   &$\beta_1$    & 0.096 & 0.730 & 0.738 & 0.956 & 0.540 & 0.008     &  & 0.090 & 0.903 & 0.923 & 0.954 & 0.822 & 0.008  \\
     &          $\beta_2$   & 0.086 & 0.710 & 0.731 & 0.968 & 0.511 &    & & 0.080 & 0.892 & 0.917 & 0.964 & 0.800 &  \\
     &          $\beta_3$   & 0.028 & 0.424 & 0.425 & 0.954 & 0.180 &     && 0.030 & 0.524 & 0.530 & 0.960 & 0.275 &       \\
   \botrule
   \end{tabular*}
 \end{table*}

\section{Simulation Studies}

\begin{figure*}[!htbp]
  \begin{center}
    \centerline{\includegraphics[width=\textwidth]{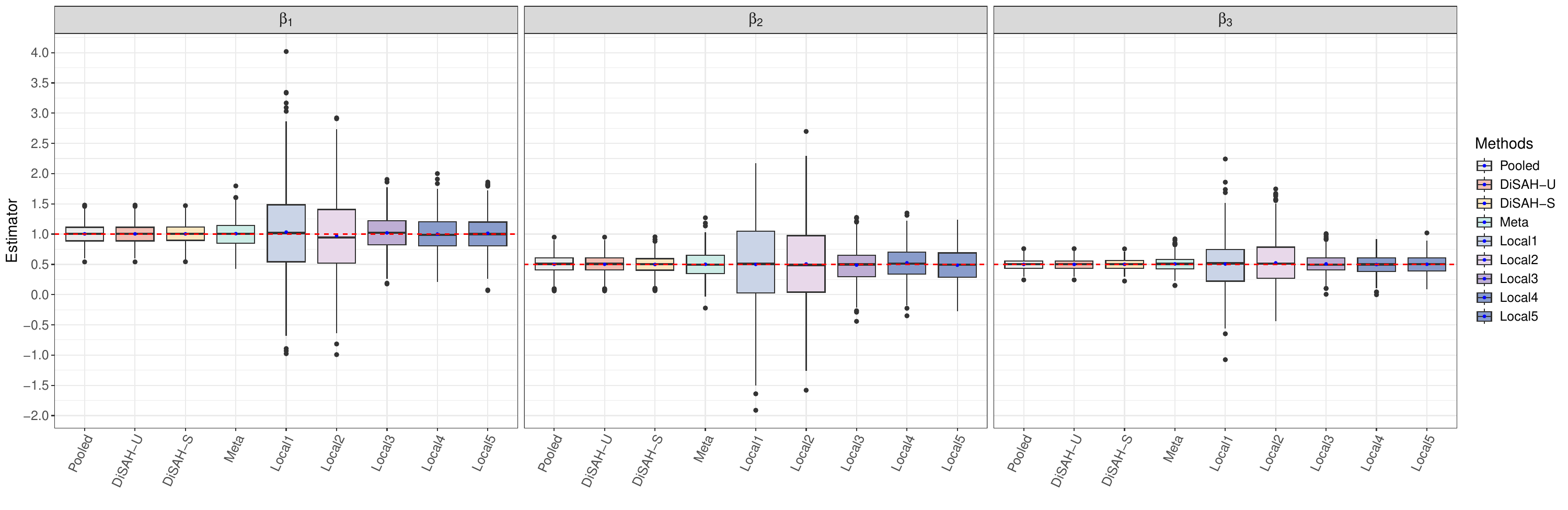}}
    \caption{Hazard difference estimates under heterogeneous baseline hazards. Box plots show the distribution of estimated hazard differences across 500 replications for five methods: DiSAH-U, DiSAH-S, Pooled, Meta, and Local. 
    Site sample sizes are $n_1=100$, $n_2=100$, $n_3=500$, $n_4=1000$, and $n_5=1000$ (configuration (ii), imbalanced).  Dashed lines indicate true parameter values.}
    \label{boxplot-sim-hete-main50}
  \end{center}
\end{figure*} 

\subsection{Simulation setup}
\label{subsec.sim}
To evaluate the empirical performance of DiSAH, we simulated distributed survival data from $K=5$ sites.
Each site generated covariates $\bX=(X_1, X_2, X_3)$ with $X_1, X_2 \overset{\mathrm{iid}}{\sim} \text{Uniform}(0,1)$ and $X_3 \overset{\mathrm{iid}}{\sim} \text{Bernoulli}(0.5)$.
The true hazard difference coefficients were $\bbeta_0=(\beta_1, \beta_2, \beta_3) = (1, 0.5, 0.5)$. 

We considered two scenarios.
In scenario 1 (homogeneous), event times $T$ were generated from the unstratified additive hazards model 
(Eq.~\eqref{m1}) with common baseline hazard $\lambda_0(t) = t^2$ across all sites. 
In scenario 2 (heterogeneous), event times were generated from the stratified additive hazards model (Eq.~\eqref{stratified_mk}) with site-specific baseline hazards: $\lambda_{01}(t) = t^2$, 
$\lambda_{02}(t) = t^3$, $\lambda_{03}(t) = t^4$, $\lambda_{04}(t) = \log(1+t) + t^3$, 
and $\lambda_{05}(t) = \log(1+t) + t^4$. 
Censoring times $C$ were generated independently from $\text{Uniform}(0.02, 1.28)$, yielding 
observed time $Y = \min(T, C)$ and event indicator $\delta = I(T \leq C)$.
For each scenario, we considered two sample size configurations: configuration (i) balanced small sites with $n_k = 100$ for all $k$ sites, and configuration (ii) imbalanced sites with $n_1 = n_2 = 100$, $n_3 = 500$, $n_4 = 1000$, and $n_5 = 1000$.

We evaluated five approaches: \textbf{Pooled}: centralized analysis using combined individual-level data (infeasible oracle); \textbf{DiSAH-U}: proposed federated estimator under the unstratified model; \textbf{DiSAH-S}: proposed federated estimator under the stratified model; \textbf{Meta}: inverse-variance weighted average of local estimates; and \textbf{Local}: site-specific estimation. All results are based on 500 replications.

\subsection{Results}

Tables~\ref{Tab.balanced} and~\ref{Tab.imbalanced} report the results for the balanced and imbalanced configurations under both scenarios, including bias, sample standard deviation (SD), average estimated standard error (SE), 95\% confidence interval coverage probability (CP), mean squared error (MSE), and computation time.
Both DiSAH-U and DiSAH-S produced nearly unbiased estimates across settings, with SD closely matching SE and coverage probabilities near the nominal 95\% level, confirming the validity of the asymptotic variance estimators.
DiSAH achieved estimation efficiency comparable to the pooled oracle, with the gap narrowing as total sample size increased. 
Both DiSAH variants substantially outperformed the Meta approach in statistical efficiency, although Meta also exhibited negligible bias. 
This occurs because DiSAH approximates the combined likelihood of the pooled data, avoiding the finite-sample bias that accumulates across individual site-specific estimators.

Figure~\ref{boxplot-sim-hete-main50} and Figures~\ref{boxplot-sim-homo-balanced},~\ref{boxplot-sim-hete-balanced} and~\ref{boxplot-sim-homo-imbalanced} in Appendix~\ref{appendix.simulation} further illustrate these findings: DiSAH estimates concentrate tightly around the true values with variability comparable to the pooled analysis, while Local estimates exhibit substantially higher variance, particularly at smaller sites. 
Between the two proposed estimators, DiSAH-S requires only one round of communication compared to three for DiSAH-U. In our simulations, this resulted in faster computation with equivalent statistical performance; in practice, the advantage would be more pronounced due to communication overhead.

\begin{figure*}[!htbp]
  \begin{center}
    \centerline{\includegraphics[width=\textwidth]{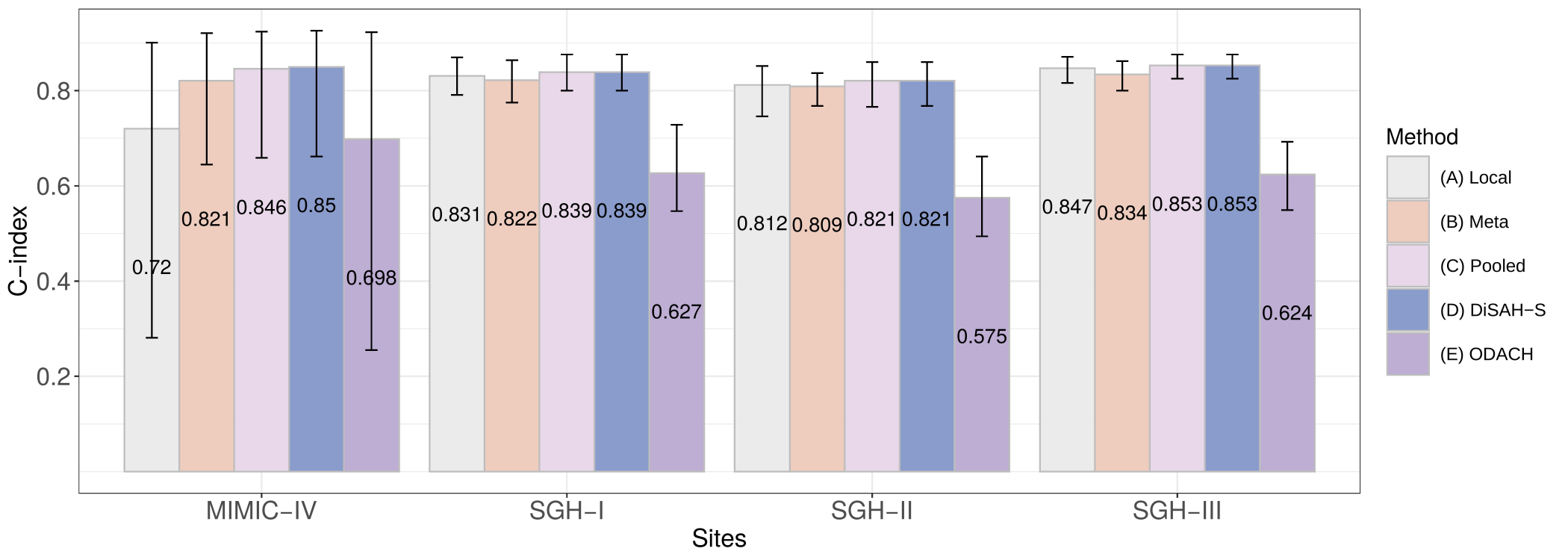}}
    \caption{Performance of Local, Meta, Pooled, DiSAH-S, and ODACH methods evaluated on MIMIC and three SGH sites. Error bars show 95\% confidence intervals.}
    \label{boxplot-realdat}
  \end{center}
\end{figure*}

\section{Real Data Application}

In-patient mortality following emergency-department (ED) admission is among the highest-stakes predictions in acute care, governing triage acuity assignment, the choice between intensive-care and general-ward disposition, and the timing of early escalation~\citep{subbe2001validation}. 
Each of these decisions turns on absolute risk, the number of additional deaths a given physiological derangement or comorbidity confers over a fixed horizon, rather than on relative risk~\citep{laupacis1988assessment}. A clinician allocating a finite number of monitored beds needs to know how many excess events a risk factor produces per patient treated, a quantity a hazard ratio cannot supply.

Yet mortality models developed at a single center transport poorly to others, because case mix, physiological baselines, comorbidity burden, and access patterns differ substantially across health systems~\citep{justice1999assessing}. 
The multi-center data that would resolve this cannot simply be pooled. Cross-national EHR governance prohibits centralizing patient records, and the protected computing environments that hold clinical data sit behind firewalls that forbid the persistent external connections iterative federated methods require~\citep{li2024federated}. For DiSAH, server-independence is therefore not a convenience but a precondition for the collaboration to exist at all.

We accordingly ask whether DiSAH can recover, from summary statistics alone and with no central server, the mortality-risk structure that a centralized analysis would identify. To do so, we use genuinely heterogeneous ED cohorts from the United States (MIMIC-IV-ED) and Singapore (Singapore General Hospital), evaluating both hazard-difference estimation and discrimination across the distributed sites.

\subsection{Data and Experimental Setup}
We assembled two heterogeneous ED cohorts: the public MIMIC-IV-ED~\citep{johnson2023mimic} dataset and private electronic health records from Singapore General Hospital (SGH)~\citep{sgh}. Further details of data cleaning and cohort formation are provided in Appendix~\ref{appendix.realdata}.
Both cohorts comprise Asian patients aged 18 years or older with complete covariate information. 
The survival outcome was 30-day in-patient mortality after ED admission. Patients who were lost to follow-up during the study period or who did not die within 30 days were treated as right-censored.

Covariates included demographics (age, gender), vital signs (pulse, respiratory rate, oxygen saturation, blood pressure), triage class, and comorbidities (myocardial infarction, congestive heart failure, peripheral vascular disease, stroke, chronic pulmonary disease, liver disease, renal disease, diabetes, malignancy). Full variable definitions are provided in Appendix~\ref{appendix.realdata}, Table~\ref{tab.descriptive}.

After preprocessing, we obtained $n = 4{,}433$ patients from MIMIC and $n = 43{,}345$ from SGH. To emulate a realistic distributed setting with site-level heterogeneity, we partitioned the SGH cohort into three non-overlapping sites (SGH-I, SGH-II, SGH-III) with proportions 30\%, 30\%, and 40\%.
This yielded four total sites: MIMIC ($n_1=4{,}433$), SGH-I ($n_2=13{,}003$), SGH-II ($n_3=13{,}004$), and SGH-III ($n_4=17{,}338$).
Table~\ref{tab.descriptive} in the Appendix summarizes descriptive statistics across sites, revealing substantial heterogeneity in outcome rates, covariate distributions, and sample sizes, ranging from 4{,}433 (MIMIC) to 17{,}338 (SGH-III).

Each site's data were split 70/30 into training and test sets. 
We benchmarked DiSAH against four comparators:
(1) \textbf{Local}, site-specific additive hazards models trained only on local data; 
(2) \textbf{Meta}, inverse-variance weighted aggregation of site-specific model coefficients;
(3) \textbf{Pooled}, centralized training on pooled multi-site data; and 
(4) \textbf{ODACH}, a federated Cox regression method~\citep{luo2022odach}\footnote{No existing federated method estimates hazard differences. ODACH is included as the closest baseline but targets hazard ratios; we therefore compare only discrimination performance.}.

\subsection{Results}
\paragraph{Hazard difference estimation.}
Table~\ref{table:RD-realdat} in Appendix~\ref{appendix.realdata} reports 
hazard difference estimates, standard errors, and $p$-values for each covariate. 
The covariate effect estimates obtained by DiSAH are very close to those obtained from pooled individual-level data, and DiSAH identified the same statistically significant risk factors as the pooled analysis. Notably, DiSAH detected risk factors that local analyses missed due to insufficient power at individual sites. For example, gender, systolic blood pressure, pulse, respiratory rate, SAO2, malignancy and myocardial infarction reached significance only under DiSAH and pooled analysis in the MIMIC cohort.
Regarding the estimation efficiency, DiSAH also achieved smaller standard errors than local analyses across all covariates, demonstrating the statistical efficiency gains from federated estimation without data sharing.

\paragraph{Prediction performance.}
Figure~\ref{boxplot-realdat} compares the C-index across methods and sites. 
DiSAH achieved discrimination comparable to pooled analysis while substantially outperforming local models, particularly on the smaller MIMIC cohort where local analysis yielded a C-index of only 0.72 with wide confidence intervals. 
The Meta approach performed between Local and DiSAH. ODACH showed notably lower discrimination, possibly due to violations of the proportional hazards assumption in this heterogeneous setting. These findings illustrate why relying solely on hazard ratio estimation may be insufficient: when the proportional hazards assumption fails to hold across heterogeneous populations, the hazard difference offers a robust and interpretable alternative.

Table~\ref{table:S4-complete-AUCt} in Appendix~\ref{appendix.realdata} further reports time-dependent AUC at $t = 7, 14, 21, 28$ days and the integrated AUC (iAUC), with 95\% confidence intervals. Results are consistent with the C-index findings: DiSAH matches pooled performance across all time horizons.

\section{Discussion}
We developed DiSAH, a communication-efficient, privacy-preserving algorithm for federated hazard difference estimation in multi-site survival data. Unlike engineering-based FL methods that require persistent server connections and iterative optimization, DiSAH is fully server-independent: the coordinator role requires only simple aggregation and can be assumed by any participating site. 
This architectural simplicity makes DiSAH particularly suitable for healthcare settings where IT infrastructure varies across institutions and protected computing environments prohibit external connections~\citep{li2024federated}.

A central finding from both our simulations and real-world application is that DiSAH achieves estimation accuracy comparable to the infeasible pooled oracle while requiring only summary-level statistics. In the homogeneous setting, the unstratified estimator (DiSAH-U) is asymptotically equivalent to the pooled estimator, a property that follows from the closed-form structure of the additive hazards estimating equation. In the heterogeneous setting, DiSAH-S accommodates site-specific baseline hazards in a single communication round, with only modest efficiency loss relative to DiSAH-U. Both variants substantially outperformed meta-analysis, which inherits the bias of local estimators, a limitation that becomes pronounced when site-specific sample sizes are small or unbalanced.

The real-world application using emergency department data from MIMIC-IV-ED and Singapore General Hospital illustrates a practical advantage of distributed estimation over local analysis. DiSAH identified risk factors for 30-day mortality, including gender, systolic blood pressure, pulse, respiratory rate, SAO2, malignancy and myocardial infarction, that individual sites lacked power to detect, while achieving discrimination comparable to pooled analysis. The notably lower performance of the federated Cox model (ODACH) in this application suggests that the proportional hazards assumption may not hold across these heterogeneous populations, reinforcing the value of additive hazards models as a complementary analytic framework.

Several limitations should be noted. First, the unstratified variant requires sharing ordered event times across sites, which, although stripped of individual identifiers, may carry residual re-identification risk in settings with very rare events. 
Second, our real-world evaluation partitioned a single SGH cohort into three pseudo-sites, which introduces sampling variability but does not fully replicate the heterogeneity arising from genuinely 
distinct institutions with different coding practices and data quality. Validation in a network of more independent clinical sites would further strengthen the evidence.

Future work includes extending DiSAH to high-dimensional covariate settings through penalized estimation, to more general censoring mechanisms such as interval-censored~\citep{wang2010regression, wang2016regression} and mixed-case~\citep{zhao2004generalized} failure time data, and developing open-source software integrated with common data models to facilitate adoption in existing distributed research networks.

\section{Conclusion}
DiSAH is a privacy-preserving, communication-efficient distributed algorithm for estimating hazard differences from multi-site survival data. Through simulations and a real-world application using cross-national emergency department records, we demonstrated that DiSAH achieves accuracy comparable to pooled analysis while substantially outperforming meta-analysis and local estimation. By providing interpretable absolute risk estimates without requiring patient-level data sharing, DiSAH offers a practical tool for collaborative survival analysis across healthcare institutions.

\backmatter







\section*{Declarations}

\subsection*{Funding}
Z.W. was supported by the Shanxi Key Laboratory of Digital Design and Manufacturing (202204010931025). S.L. was supported by the Khoo Postdoctoral Fellowship Award (Duke-NUS-KPFA/2025/0081), funded by the Estate of Tan Sri Khoo Teck Puat. M.E.H.O. was  supported by the National Medical Research Council, Singapore, through Clinician Scientist Awards (NMRC/CSA/024/2010 and NMRC/CSA/0049/2013), and the Ministry of Health, Singapore, through the Health Services Research Grant (HSRG/0021/2012). The  funders had no role in study design, data collection, analysis, interpretation, or manuscript preparation.

\subsection*{Conflict of interest}
The authors declare that they have no competing interests.

\subsection*{Data and Code availability}
Code to reproduce all simulation studies and implement the proposed methods is available at \url{https://github.com/siqili0325/DiSAH}. 
The MIMIC-IV-ED dataset is publicly available through PhysioNet (\url{https://physionet.org/content/mimic-iv-ed/}) after credentialing. The SGH dataset is confidential and cannot be shared publicly due to institutional data governance policies.

\subsection*{Author contribution}
Z.W. developed the methodology, derived the theoretical results, designed the algorithms, conducted the simulation studies, and drafted the manuscript. S.L. conducted the real-world data experiments, performed the data analysis, and drafted the manuscript. M.E.H.O. provided the SGH clinical data and clinical domain expertise. N.L. conceived the study. All authors revised and approved the final manuscript.


\bibliography{sn-bibliography}

\newpage
\begin{appendices}

\section{Theoretical Guarantees}\label{sec.thm}

We establish asymptotic properties of the DiSAH estimators under the following regularity conditions, which are standard in additive risk models and ensure the validity of the law of large numbers and martingale central limit theorem~\citep{lin1994, Kulich2000}.

\noindent\textbf{C1.}
The subjects are independent across $(i,k)$, and within each stratum $k$, $\{(N_{ik}(t),Y_{ik}(t),X_{ik})\}$ are identically distributed.\\
\noindent\textbf{C2.}
The covariate process $X$ is time-independent and uniformly bounded, i.e., $\|X\|\le C$ almost surely for some finite constant $C$.\\
\noindent\textbf{C3.}
The at-risk indicator $Y(t)$ is left-continuous with right limits and satisfies $P\{Y(t)=1|X\}>0$ for all $t\in[0,\tau]$, where $\tau$ satisfies $P(Y\geq\tau)>0$.\\
\noindent\textbf{C4.}
The baseline hazard functions $\lambda_0(t)$ and $\lambda_{0k}(t)$ are nonnegative and locally integrable on $[0,\tau]$.\\
\noindent\textbf{C5.}
The matrix $A=\int_0^\tau E\!\left[Y(t)\{X-\mu(t)\}^{\otimes2}\right]dt$ is positive definite.\\
\noindent\textbf{C6.}
For each site $k$, the matrix
$A_k=\int_0^\tau E\!\left[Y_{ik}(t)\{X_{ik}-\mu_k(t)\}^{\otimes2}\right]dt$
is positive definite.

\begin{theorem}[Asymptotic Normality of DiSAH-U]
Under the unstratified additive hazards model and above conditions, let $\hat\bbeta_{U}$ be the solution to the estimating equation~\eqref{m4}, with above conditions C1--C6, we have: 
\[
\sqrt{n}(\hat\bbeta_{U}-\bbeta_0)
\;\xrightarrow{d}\;
N\!\left(0,\,A^{-1}\Sigma A^{-1}\right),
\]
where
\[A=\int_0^\tau E\!\left[ Y(t)\{\bX-\mu(t)\}^{\otimes 2} \right]dt, \] and
\[ \Sigma = E\!\left[\int_0^\tau \{\bX-\mu(t)\}^{\otimes 2} dN(t) \right]\] with $\mu(t)=E(\bX\mid Y(t)=1)$.
\end{theorem}
The covariance matrix of $\hat\bbeta_{U}$ can be estimated by the sandwich formula
\[\hat{cov}(\hat\bbeta_{U}) = \hat{A}^{-1}\hat{\Sigma}\hat{A}^{-1},\]
where a consistent estimator for $A$ under the unstratified additive risks model is $\hat{A} =n^{-1}\sum_{k=1}^{K}\sum_{l=1}^{n_{k}}\sum_{i=1}^{ n}I(y_{lk}\geq y_{(i)})\big\{ \bx_{lk}-\bar{\bx}(y_{(i)}) \big\}^{\otimes 2}\Delta y_{(i)}$ and $\Sigma$ can be consistently estimated by $\hat{\Sigma} = n^{-1}\sum_{k=1}^{K}\sum_{i=1}^{n_{k}}\delta_{ik}\big\{ \bx_{ik}-\bar{\bx}(y_{ik}) \big\}^{\otimes 2}.$

Consistency of $\hat{A} $ and $\hat{\Sigma} $ follows from the law of large numbers, together with uniform convergence of $\bar{\bx}(t)$ to $\mu(t)$. The proof is provided in Appendix~\ref{append.proof.thm1}.

In summary, the proposed distributed estimator for DiSAH-U is asymptotically equivalent to the pooled estimator because both are solutions to the same estimating equation, differing only in computational implementation rather than statistical formulation.

\begin{theorem}[Asymptotic Normality of DiSAH-S]
Under the stratified additive hazards model, let $\hat\bbeta_{S}$ be the solution to the estimating equation \eqref{m6}.
Suppose that $n_k/n\to p_k>0$. Under Conditions C1--C6, we have
\[\sqrt{n}(\hat\bbeta_{S}-\bbeta_0)\;\xrightarrow{d}\;N\!\left(0,\, A_{S}^{-1}\Sigma_{S}A_{S}^{-1}\right),\]
where
\[A_{S}= \sum_{k=1}^K p_k \int_0^\tau E\!\left[ Y_{ik}(t)\{\bX_{ik}-\mu_k(t)\}^{\otimes 2} \right]dt, \] and
\[\Sigma_{S} = \sum_{k=1}^K p_k E\!\left[ \int_0^\tau \{\bX_{ik}-\mu_k(t)\}^{\otimes 2}\,dN_{ik}(t)  \right]\]
with $\mu_k(t)=E(\bX_{ik}\mid Y_{ik}(t)=1)$.
\end{theorem}
The covariance matrix of $\hat\bbeta_{S}$ can be estimated by the sandwich formula
$\hat{cov}(\hat\bbeta_{S}) = \hat{A}_{S}^{-1}\hat{\Sigma}_{S}\hat{A}_{S}^{-1}$,
where $\hat{A}_{S} = \sum_{k=1}^K p_k  A_k$ and $\hat{\Sigma}_{S} = \sum_{k=1}^K p_k \Sigma_k .$ The proof is provided in Appendix~\ref{append.proof.thm2}.

\section{Theoretical Proofs}\label{sec.proofs}

\subsection{Proof of Theorem 1}
\label{append.proof.thm1}

\begin{proof}
We assume the unstratified additive hazards model that all subjects share a common baseline hazard function $\lambda_0(t)$.
By the Doob--Meyer decomposition,
\[
dN_{ik}(t) = Y_{ik}(t)\{\lambda_0(t)+\beta^\top X_{ik}\}\,dt + dM_{ik}(t),
\]
where $M_{ik}(t)$ is a martingale with respect to the filtration generated by the observed history.

Substituting the Doob--Meyer decomposition into $U(\beta)$ (\eqref{m4}) and evaluating at the true value $\beta_0$, we obtain
\[
\begin{aligned}
U(\beta_0) &= \sum_{k=1}^K \sum_{i=1}^{n_k}\int_0^\tau \{X_{ik}-\bar X(t)\} \{dM_{ik}(t)+Y_{ik}(t)\lambda_0(t)\,dt\}.
\end{aligned}
\]
Since
\[\sum_{k=1}^K \sum_{i=1}^{n_k} \{X_{ik}-\bar X(t)\}Y_{ik}(t)=0,\]
then
\begin{equation}\label{eq:A1}
U(\beta_0) = \sum_{k=1}^K \sum_{i=1}^{n_k} \int_0^\tau \{X_{ik}-\bar X(t)\}\, dM_{ik}(t).
\end{equation}

Let $\hat\beta$ denote the solution to $U(\beta)=0$.
A Taylor expansion around $\beta_0$ gives
\begin{equation}\label{eq:A2}
0 = U(\beta_0) + \left.\frac{\partial U(\beta)}{\partial\beta}\right|_{\beta=\beta_0}
(\hat\beta-\beta_0) + o_p(\sqrt n\|\hat\beta-\beta_0\|).
\end{equation}
Under standard regularity conditions,
\begin{equation}\label{eq:A3}
-\frac{1}{n}\frac{\partial U(\beta)}{\partial\beta} \xrightarrow{p}
A=\int_0^\tau E\!\left[ Y(t)\{X-\mu(t)\}^{\otimes 2} \right]dt,
\end{equation}
where $\mu(t)=E(X\mid Y(t)=1)$ and $A$ is nonsingular.

By the martingale central limit theorem applied to \eqref{eq:A1}, we next derive
\begin{equation}\label{eq:A4}
n^{-1/2}U(\beta_0) \xrightarrow{d} N(0,\Sigma),
\end{equation}
where
\[\Sigma = E\!\left[\int_0^\tau \{X-\mu(t)\}^{\otimes 2} dN(t) \right].\]

Define
\[U_n(t) = \sum_{k=1}^K\sum_{i=1}^{n_k} \int_0^t \{X_{ik}-\bar X(s)\}\,dM_{ik}(s).\]

Because the martingales $\{M_{ik}(t)\}$ are orthogonal across $(i,k)$,
the predictable quadratic variation of $U_n(t)$ is
\[
\begin{aligned}
\langle U_n\rangle(t)
&= \sum_{k=1}^K\sum_{i=1}^{n_k} \int_0^t \{X_{ik}-\bar X(s)\}^{\otimes 2}\,d\langle M_{ik}\rangle(s).
\end{aligned}
\]
For the unstratified additive hazards model,
\[d\langle M_{ik}\rangle(s) = Y_{ik}(s)\{\lambda_0(s)+\beta_0^\top X_{ik}\}\,ds.\]
Hence,
\[
\begin{aligned}
\langle U_n\rangle(\tau)
&= \sum_{k=1}^K\sum_{i=1}^{n_k} \int_0^\tau \{X_{ik}-\bar X(s)\}^{\otimes 2} Y_{ik}(s)\{\lambda_0(s)+\beta_0^\top X_{ik}\}\,ds.
\end{aligned}
\]

By the uniform law of large numbers for predictable processes,
\[\bar X(t)\xrightarrow{p}\mu(t)\quad \text{uniformly in } t\in[0,\tau],\]
where $\mu(t)=E(X\mid Y(t)=1)$.
Therefore,
\[\frac{1}{n}\langle U_n\rangle(\tau) \xrightarrow{p} \Sigma,\]
with
\[\Sigma = E\!\left[\int_0^\tau \{X-\mu(t)\}^{\otimes 2} Y(t)\{\lambda_0(t)+\beta_0^\top X\} \,dt \right]
= E\!\left[\int_0^\tau \{X-\mu(t)\}^{\otimes 2} dN(t) \right].\]
By the martingale central limit theorem,
\[
n^{-1/2}U(\beta_0) \xrightarrow{d} N(0,\Sigma).
\]

Consequently, combining the above results \eqref{eq:A2}, \eqref{eq:A3} and \eqref{eq:A4}, we can obtain
\[\sqrt n(\hat\beta-\beta_0) \xrightarrow{d} N\!\left(0,\ A^{-1}\Sigma A^{-1}\right).\]

\paragraph{Estimation of covariance matrix.}
In the unstratified additive hazards model, the predictable variation of the counting process satisfies
\[E\!\left\{ dN_{ik}(t) \mid \mathcal{F}_{t^-} \right\}
= Y_{ik}(t) \left\{ \lambda_{0}(t) + \beta_0^{\top} X_{ik}\right\}\,dt.\]
Therefore, a consistent estimator of $\Sigma$ is obtained by directly replacing the intensity term with the observed counting process increments.
\[\hat{\Sigma}
=n^{-1}\sum_{k=1}^{K} \sum_{i=1}^{n_k}  \delta_{ik}\left\{ X_{ik} - \bar{X}(y_{i}) \right\}^{\otimes 2}.
\]

\end{proof}

\subsection{Proof of Theorem 2}
\label{append.proof.thm2}

\begin{proof}
In the stratified additive hazards model, by the Doob--Meyer decomposition,
\[dN_{ik}(t)= Y_{ik}(t)\{\lambda_{0k}(t)+\beta^\top X_{ik}\}\,dt + dM_{ik}(t),\]
where $\{M_{ik}(t)\}$ are martingales with respect to the filtration $\mathcal{F}_t$.

Consider the stratified estimating function (\eqref{m6}). Evaluating at $\beta_0$ yields
\[
\begin{aligned}
S(\beta_0)&=\sum_{k=1}^K \sum_{i=1}^{n_k}\int_0^\tau\{X_{ik}-\bar X_k(t)\}\{dM_{ik}(t)+Y_{ik}(t)\lambda_{0k}(t)\,dt\}.
\end{aligned}
\]
For each site $k$,
\[
\sum_{i=1}^{n_k}\{X_{ik}-\bar X_k(t)\}Y_{ik}(t)=0,
\]
and hence
\begin{equation}
S(\beta_0)=\sum_{k=1}^K \sum_{i=1}^{n_k}\int_0^\tau\{X_{ik}-\bar X_k(t)\}\, dM_{ik}(t).
\label{eq:A5}
\end{equation}

Let $\hat\beta_S$ satisfy $S(\hat\beta)=0$. A Taylor expansion gives
\begin{equation}
0=S(\beta_0)+\left.\frac{\partial S(\beta)}{\partial\beta}\right|_{\beta=\beta_0}
(\hat\beta_{S}-\beta_0) + o_p(\sqrt n\|\hat\beta_{S}-\beta_0\|).
\label{eq:A6}
\end{equation}
Assume $n_k/n \to p_k>0$. Then
\begin{equation}
-\frac{1}{n}\frac{\partial S(\beta)}{\partial\beta}\;\xrightarrow{p}A_{S}=\sum_{k=1}^{K}p_k\int_0^\tau
E\!\left[Y_{ik}(t)\{X_{ik}-\mu_k(t)\}^{\otimes 2}\right]dt,
\label{eq:A7}
\end{equation}
where $\mu_k(t)=E(X_{ik}\mid Y_{ik}(t)=1)$.

From \eqref{eq:A6} and independence across sites, we next show
\begin{equation}
n^{-1/2}S(\beta_0)\xrightarrow{d}N(0,\Sigma_{S}),
\label{eq:A8}
\end{equation}
with
\[
\Sigma_{S}=\sum_{k=1}^K p_kE\!\left[\int_0^\tau\{X_{ik}-\mu_k(t)\}^{\otimes 2}dN_{ik}(t)\right].
\]

Define the site-specific score process
\[
S_k(\beta_0)=\sum_{i=1}^{n_k}\int_0^\tau\{X_{ik}-\bar X_k(t)\}\, dM_{ik}(t).
\]
Under the assumption that different sites are independent,
$\{S_k(\beta_0)\}_{k=1}^K$ are independent martingales.
Therefore,
\[
n^{-1/2}S(\beta_0)=\sum_{k=1}^K \sqrt{\frac{n_k}{n}}\left\{ n_k^{-1/2} S_k(\beta_0) \right\}.
\]

We next derive the asymptotic covariance of $n_k^{-1/2}S_k(\beta_0)$.

For each subject $(i,k)$, the counting process $N_{ik}(t)$ admits the Doob--Meyer decomposition
\[
dN_{ik}(t)=Y_{ik}(t)\{\lambda_{0k}(t)+\beta_0^\top X_{ik}\}\,dt+dM_{ik}(t),
\]
with predictable variation
\[
d\langle M_{ik}\rangle(t)=Y_{ik}(t)\{\lambda_{0k}(t)+\beta_0^\top X_{ik}\}\,dt.
\]
Since the martingales $\{M_{ik}(t)\}_{i=1}^{n_k}$ are orthogonal,
the predictable quadratic variation of $S_k(\beta_0)$ is given by
\[
\begin{aligned}
\langle S_k(\beta_0)\rangle
&=\sum_{i=1}^{n_k}\int_0^\tau\{X_{ik}-\bar X_k(t)\}^{\otimes 2}\, d\langle M_{ik}\rangle(t) \\
&=\sum_{i=1}^{n_k}\int_0^\tau\{X_{ik}-\bar X_k(t)\}^{\otimes 2}Y_{ik}(t)\{\lambda_{0k}(t)+\beta_0^\top X_{ik}\}\,dt.
\end{aligned}
\]

By the uniform law of large numbers for predictable processes,
\[
\bar X_k(t)\;\xrightarrow{p}\;\mu_k(t)=E(X_{ik}\mid Y_{ik}(t)=1),
\]
uniformly in $t\in[0,\tau]$.
Consequently,
\[
\frac{1}{n_k}\langle S_k(\beta_0)\rangle\;\xrightarrow{p}\;\Sigma_k,
\]
where
\[
\begin{aligned}
\Sigma_k
&=E\!\left[\int_0^\tau\{X_{ik}-\mu_k(t)\}^{\otimes 2}Y_{ik}(t)\{\lambda_{0k}(t)+\beta_0^\top X_{ik}\}\,dt\right] \\
&=E\!\left[\int_0^\tau\{X_{ik}-\mu_k(t)\}^{\otimes 2}Y_{ik}(t)dN_{ik}(t)\right].
\end{aligned}
\]

Assume that $n_k/n \to p_k>0$ for each $k$.
Then the asymptotic covariance of $n^{-1/2}S(\beta_0)$ is
\[
\Sigma_{S}=\sum_{k=1}^K p_k \Sigma_k
= \sum_{k=1}^K p_k E\!\left[ \int_0^\tau \{X_{ik}-\mu_k(t)\}^{\otimes 2}dN_{ik}(t)\right].
\]
By the martingale central limit theorem,
\[
n^{-1/2}S(\beta_0)\xrightarrow{d} N(0,\Sigma_{S}).
\]

Therefore, combining the above results \eqref{eq:A6}, \eqref{eq:A7} and \eqref{eq:A8}, we can obtain
\[
\sqrt n(\hat\beta-\beta_0)\xrightarrow{d}N\!\left(0,\ A_{S}^{-1}\Sigma_{S}A_{S}^{-1}\right).
\]

\paragraph{Estimation of covariance matrix.}
In the stratified additive hazards model, the predictable variation of the counting process satisfies
\[E\!\left\{ dN_{ik}(t) \mid \mathcal{F}_{t^-} \right\}
= Y_{ik}(t) \left\{ \lambda_{0k}(t) + \beta_0^{\top} X_{ik}\right\}\,dt.\]
Therefore, a consistent estimator of $\Sigma_S$ is obtained by directly replacing the intensity term with the observed counting process increments.
\[\hat{\Sigma}_{S}
=\sum_{k=1}^{K} p_k n_{k}^{-1}\sum_{i=1}^{n_k}  \delta_{ik}\left\{ X_{ik} - \bar{X}_{k}(y_{i}) \right\}^{\otimes 2}.
\]

\end{proof}

\section{Experiment Details}\label{sec.experiments}

\subsection{Simulation Studies}\label{appendix.simulation}

Extra simulation results are available in Table~\ref{Tab.imbalanced}.

\begin{table*}[!htbp]
  \caption{Simulation results for the hazard difference estimation when $n_1=n_2 =100$, $n_3=500$, $n_4= 1000$, and $n_5=1000$ (configuration (ii), imbalanced), with computation time in seconds.}\label{Tab.imbalanced}
   \setlength{\tabcolsep}{0.06cm}
   \centering
   \footnotesize
   \begin{tabular*}{\textwidth}{@{\extracolsep\fill}lcccccccccccccc@{\extracolsep\fill}}
   \toprule
   & &  \multicolumn{6}{@{}c@{}}{Scenario 1 (homogeneous)} & & \multicolumn{6}{@{}c@{}}{Scenario 2 (heterogeneous)}\\
   \cmidrule{3-8} \cmidrule{10-15}
   Methods&Para&Bias&SD&SE&CP&MSE&Runtime& &Bias &SD &SE &CP&MSE&Runtime\\
   \midrule
   \addlinespace
   Pooled   &$\beta_1$  & 0.004 & 0.134 & 0.136 & 0.954 & 0.018 & 0.902    &   & 0.004 & 0.157 & 0.157 & 0.952 & 0.025 & 0.904 \\
             &$\beta_2$   & $-$0.001 & 0.130 & 0.135 & 0.962 & 0.017 &             &  & 0.002 & 0.151 & 0.156 & 0.950 & 0.023 & \\
             & $\beta_3$   & 0.000 & 0.077 & 0.079 & 0.966 & 0.006 &             &  & $-$0.001 & 0.086 & 0.091 & 0.966 & 0.007 &  \\
   \addlinespace
   DiSAH-U    &$\beta_1$ & 0.004 & 0.134 & 0.136 & 0.954 & 0.018 & 6.270   &   & 0.004 & 0.157 & 0.157 & 0.952 & 0.025 & 6.254 \\
                       & $\beta_2$   & $-$0.001 & 0.130 & 0.135 & 0.962 & 0.017 &             &  & 0.002 & 0.151 & 0.156 & 0.950 & 0.023 &  \\
                        & $\beta_3$    & 0.000 & 0.077 & 0.079 & 0.966 & 0.006 &             &  & $-$0.001 & 0.086 & 0.091 & 0.966 & 0.007 &    \\
   \addlinespace
   DiSAH-S  &$\beta_1$    & 0.004 & 0.136 & 0.136 & 0.950 & 0.018 & 0.440   &   & 0.004 & 0.160 & 0.157 & 0.940 & 0.025 & 0.442 \\
                      & $\beta_2$    & $-$0.001 & 0.130 & 0.135 & 0.962 & 0.017 &             &  & $-$0.003 & 0.151 & 0.156 & 0.956 & 0.023 & \\
                      &$\beta_3$     & 0.001 & 0.077 & 0.079 & 0.968 & 0.006 &             &  & 0.000 & 0.086 & 0.091 & 0.970 & 0.007 &   \\
   \addlinespace
  Meta    &$\beta_1$   & 0.007 & 0.229 & 0.223 & 0.952 & 0.052 & 0.353   &   & 0.008 & 0.225 & 0.226 & 0.952 & 0.050 & 0.352 \\
            &  $\beta_2$   & 0.001 & 0.227 & 0.221 & 0.946 & 0.052 &     & & 0.001 & 0.219 & 0.223 & 0.960 & 0.048 &  \\
            &  $\beta_3$   & 0.009 & 0.131 & 0.129 & 0.958 & 0.017 &   & & 0.007 & 0.125 & 0.130 & 0.962 & 0.016 &  \\
   \addlinespace
 Local 1   &$\beta_1$   & 0.034 & 0.746 & 0.739 & 0.952 & 0.556 & 0.009   &     & 0.034 & 0.746 & 0.739 & 0.952 & 0.556 & 0.009  \\
   &          $\beta_2$    & 0.000 & 0.720 & 0.725 & 0.962 & 0.517 &     &   & 0.000 & 0.720 & 0.725 & 0.962 & 0.517 &  \\
   &          $\beta_3$    & 0.006 & 0.426 & 0.420 & 0.946 & 0.181 &   &   & 0.006 & 0.426 & 0.420 & 0.946 & 0.181 &    \\
   \addlinespace
 Local 2   &$\beta_1$  & $-$0.035 & 0.717 & 0.732 & 0.964 & 0.515 & 0.008    &   & $-$0.029 & 0.656 & 0.672 & 0.952 & 0.431 & 0.008 \\
   &          $\beta_2$  & 0.006 & 0.725 & 0.726 & 0.956 & 0.525 &    & & 0.006 & 0.673 & 0.667 & 0.954 & 0.451 & \\
   &          $\beta_3$  & 0.028 & 0.425 & 0.424 & 0.960 & 0.181 &   &  & 0.024 & 0.387 & 0.390 & 0.956 & 0.150 &    \\
   \addlinespace
 Local 3   &$\beta_1$  & 0.019& 0.323 & 0.316 & 0.942 & 0.105 & 0.051   & & 0.020 & 0.291 & 0.281 & 0.950 & 0.085& 0.049  \\
   &          $\beta_2$  & $-$0.008 & 0.309 & 0.315 & 0.952 & 0.095 &     && $-$0.012 & 0.278 & 0.280 & 0.952 & 0.078 &\\
   &          $\beta_3$  & 0.011 & 0.186 & 0.184 & 0.938 & 0.035 &   &   & 0.008 & 0.163 & 0.163 & 0.950 & 0.026 &   \\
   \addlinespace
 Local 4   &$\beta_1$  & 0.001 & 0.224 & 0.224 & 0.950 & 0.050 & 0.143 & & 0.000 & 0.292 & 0.287 & 0.954 & 0.085 & 0.143 \\
   &          $\beta_2$   & 0.024 & 0.210 & 0.222 & 0.952 & 0.044 &    & & 0.025 & 0.273& 0.285 & 0.956 & 0.075 &  \\
   &          $\beta_3$   & $-$0.001 & 0.123 & 0.130 & 0.956 & 0.015 &    & &$-$0.003 & 0.157 & 0.166 & 0.968 & 0.025 &    \\
   \addlinespace
   Local 5   &$\beta_1$    & 0.014 & 0.233 & 0.224 & 0.936 & 0.054 & 0.144     &  & 0.014 & 0.290 & 0.279 & 0.938 & 0.084 & 0.143  \\
     &          $\beta_2$   & $-$0.017 & 0.223 & 0.223 & 0.938 & 0.050 &    & & $-$0.014& 0.281 & 0.278 & 0.946 & 0.079 &  \\
     &          $\beta_3$   & $-$0.001 & 0.131 & 0.130 & 0.950 & 0.017 &     && 0.002 & 0.162 & 0.161 & 0.932 & 0.026 &       \\
   \botrule
   \end{tabular*}
 \end{table*}

\begin{figure*}[!htbp]
  \begin{center}
    \centerline{\includegraphics[width=\textwidth]{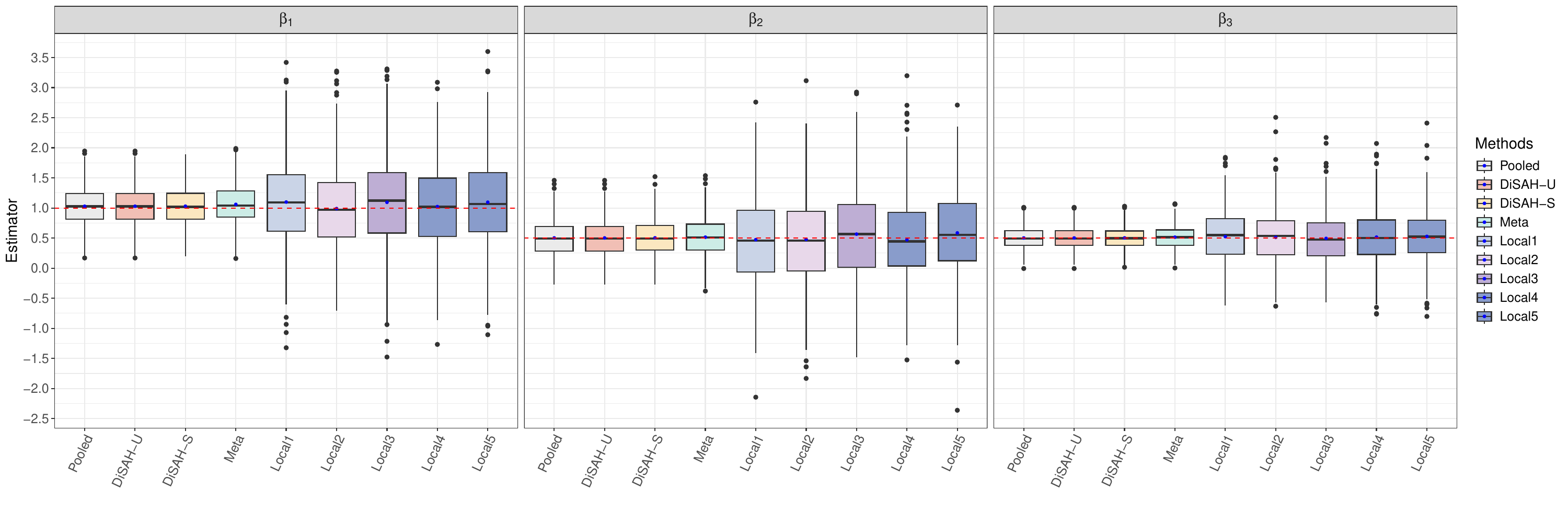}}
    \caption{
    Survival hazard difference estimates under homogeneous baseline hazards. Box plots show the distribution of estimated hazard differences across 500 replications for five methods: DiSAH-U, DiSAH-S, Pooled, Meta, and Local. 
    Site sample sizes are $n_{1}=n_{2}=n_{3}=n_{4}=n_{5}=100 $ (configuration (i), balanced).  Dashed lines indicate true parameter values.
    }
    \label{boxplot-sim-homo-balanced}
  \end{center}
\end{figure*}

\begin{figure*}[!htbp]
  \begin{center}
    \centerline{\includegraphics[width=\textwidth]{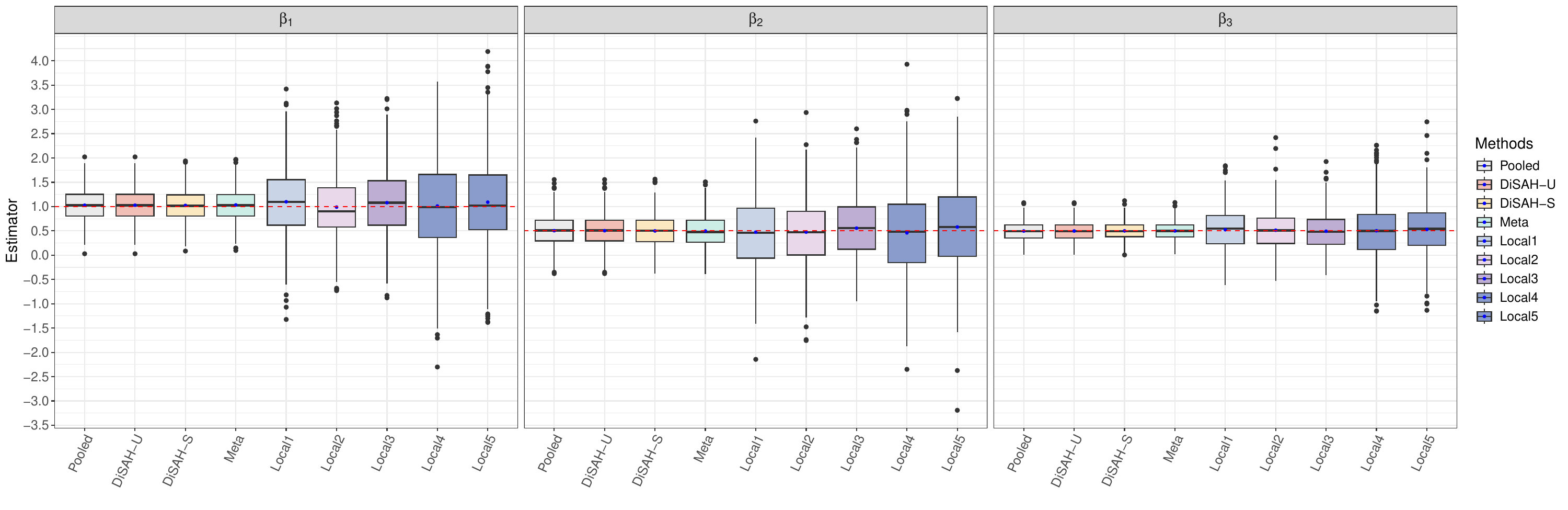}}
    \caption{
    Survival hazard difference estimates under heterogeneous baseline hazards. Box plots show the distribution of estimated hazard differences across 500 replications for five methods: DiSAH-U, DiSAH-S, Pooled, Meta, and Local. 
    Site sample sizes are $n_{1}=n_{2}=n_{3}=n_{4}=n_{5}=100 $ (configuration (i), balanced).  Dashed lines indicate true parameter values.}
    \label{boxplot-sim-hete-balanced}
  \end{center}
\end{figure*}

\begin{figure*}[!htbp]
  \begin{center}
    \centerline{\includegraphics[width=\textwidth]{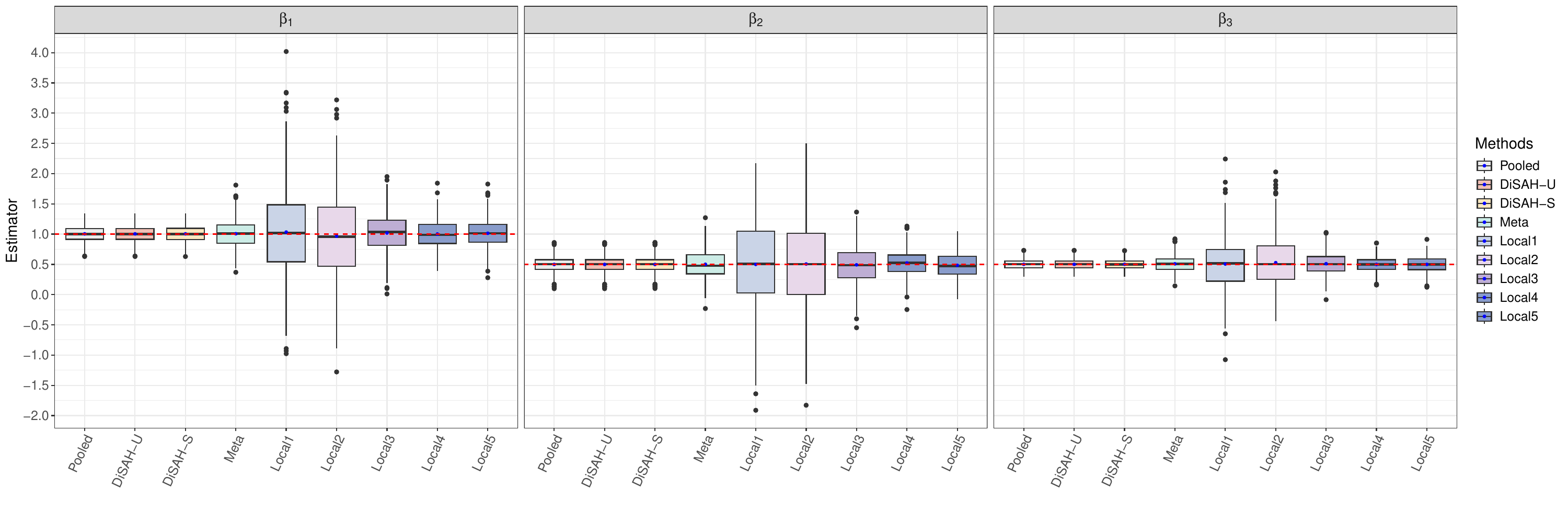}}
    \caption{Survival hazard difference estimates under homogeneous baseline hazards. Box plots show the distribution of estimated hazard differences across 500 replications for five methods: DiSAH-U, DiSAH-S, Pooled, Meta, and Local. 
    Site sample sizes are $n_1=100$, $n_2=100$, $n_3=500$, $n_4=1000$, and $n_5=1000$ (configuration (ii), imbalanced).  Dashed lines indicate true parameter values.}
    \label{boxplot-sim-homo-imbalanced}
  \end{center}
\end{figure*}

\subsection{Real Data}\label{appendix.realdata}

We utilized the MIMIC-IV Emergency Department dataset~\citep{johnson2023mimic} following the data cleaning procedure in~\citep{Xie2022} and electronic health records collected from Singapore General Hospital~\citep{sgh} for real-world data evaluation. The EHR data of SGH was extracted from the SingHealth Electronic Health Intelligence System. The collection and retrospective analysis of the EHR data were conducted under a waiver of informed consent. 
The data we analyzed has been de-identified to protect patient privacy while enabling research use.

\begin{table*}[!htbp]
\centering
\caption{Characteristics of study participants for four sites, MIMIC, SGH-I, SGH-II, and SGH-III. Continuous variables are presented as Mean (SD); binary/categorical variables are presented as Count (\%).}
\label{tab.descriptive}
\begin{center}
\footnotesize
\resizebox{\textwidth}{!}{
\begin{tabular}{lccccc}
\toprule
\textbf{Characteristic} & \textbf{MIMIC} & \textbf{SGH-I} & \textbf{SGH-II} & \textbf{SGH-III} & \textbf{P-value} \\ 
\midrule
No.\ of participants & 4433 & 13{,}003 & 13{,}004 & 17{,}338 & \\
Age, years (mean (SD)) & 52.36 (21.56) & 62.96 (18.27) & 62.88 (18.43) & 62.95 (18.28) & $<0.001$ \\
Gender (\%) & & & & & $<0.001$ \\
\quad Male & 2048 (46.2) & 6757 (52.0) & 6847 (52.7) & 9021 (52.0) & \\
\quad Female & 2385 (53.8) & 6246 (48.0) & 6157 (47.3) & 8317 (48.0) & \\
Triage class (\%) & & & & & $<0.001$ \\
\quad P1 & 271 (6.1) & 1910 (14.7) & 1899 (14.6) & 2580 (14.9) & \\
\quad P2 & 2126 (48.0) & 8992 (69.2) & 9018 (69.3) & 11{,}969 (69.0) & \\
\quad P3 \& P4 & 2036 (45.9) & 2101 (16.1) & 2087 (16.1) & 2789 (16.1) & \\
Diastolic BP (mean (SD)) & 75.72 (15.27) & 68.98 (15.54) & 68.87 (15.17) & 68.85 (15.47) & $<0.001$ \\
Systolic BP (mean (SD)) & 131.77 (22.77) & 128.34 (21.98) & 127.99 (21.80) & 127.95 (21.68) & $<0.001$ \\
Pulse, bpm (mean (SD)) & 85.97 (18.80) & 79.86 (17.68) & 79.85 (17.97) & 80.04 (18.01) & $<0.001$ \\
Resp.\ rate (mean (SD)) & 17.46 (2.35) & 17.29 (1.53) & 17.29 (1.45) & 17.29 (1.51) & $<0.001$ \\
SAO2 (mean (SD)) & 98.47 (2.84) & 97.72 (5.01) & 97.83 (4.04) & 97.83 (4.31) & $<0.001$ \\
MI (\%) & 43 (1.0) & 770 (5.9) & 756 (5.8) & 1029 (5.9) & $<0.001$ \\
CHF (\%) & 77 (1.7) & 1004 (7.7) & 1001 (7.7) & 1321 (7.6) & $<0.001$ \\
PVD (\%) & 59 (1.3) & 610 (4.7) & 571 (4.4) & 739 (4.3) & $<0.001$ \\
Stroke (\%) & 50 (1.1) & 1562 (12.0) & 1583 (12.2) & 2015 (11.6) & $<0.001$ \\
Pulmonary (\%) & 115 (2.6) & 941 (7.2) & 910 (7.0) & 1209 (7.0) & $<0.001$ \\
Renal (\%) & 114 (2.6) & 2954 (22.7) & 2991 (23.0) & 3973 (22.9) & $<0.001$ \\
Malignancy (\%) & & & & & $<0.001$ \\
\quad None & 4130 (93.1) & 10{,}687 (82.2) & 10{,}716 (82.4) & 14{,}395 (83.0) & \\
\quad Local tumor/leukemia & 225 (5.1) & 1075 (8.3) & 1111 (8.5) & 1391 (8.0) & \\
\quad Metastatic solid tumor & 78 (1.8) & 1241 (9.5) & 1177 (9.1) & 1552 (9.0) & \\
Liver disease (\%) & & & & & $<0.001$ \\
\quad None & 4305 (97.1) & 12{,}113 (93.2) & 12{,}088 (93.0) & 16{,}117 (93.0) & \\
\quad Mild & 106 (2.4) & 628 (4.8) & 682 (5.2) & 883 (5.1) & \\
\quad Severe & 22 (0.5) & 262 (2.0) & 234 (1.8) & 338 (1.9) & \\
Diabetes (\%) & & & & & $<0.001$ \\
\quad None & 4216 (94.7) & 8589 (66.0) & 8532 (65.6) & 11{,}395 (65.7) & \\
\quad w/o complications & 201 (4.5) & 347 (2.7) & 349 (2.7) & 486 (2.8) & \\
\quad w/ complications & 35 (0.8) & 4067 (31.3) & 4123 (31.7) & 5457 (31.5) & \\
Survival time (mean (SD)) & 3.4 (4.86) & 6.66 (6.93) & 6.61 (6.89) & 6.73 (7.01) & $<0.001$ \\
Status (\%) & & & & & $<0.001$ \\
\quad True & 39 (0.88) & 349 (2.68) & 328 (2.52) & 471 (2.72) & \\
\quad False & 4394 (99.12) & 12{,}654 (97.32) & 12{,}676 (97.48) & 16{,}867 (97.28) & \\
\botrule
\end{tabular}
}
\end{center}
\end{table*}

\begin{landscape}
\begin{table*}[!htbp]
\caption{Hazard difference estimates for 30-day mortality by covariate. Point estimates, standard errors, and $p$-values are shown for local sites, DiSAH, and pooled analysis. DiSAH identifies significant risk factors consistent with the pooled oracle while achieving smaller standard errors than local analyses.}
\label{table:RD-realdat}
\centering
\footnotesize
\begin{tabular}{lcccccc}
\toprule
\textbf{Variables} & \textbf{MIMIC} & \textbf{SGH-I} & \textbf{SGH-II} & \textbf{SGH-III} & \textbf{Pooled} & \textbf{DiSAH} \\
\midrule
Age & $9.16 \times 10^{-5}$ & $1.05 \times 10^{-4}$ & $9.64 \times 10^{-5}$ & $8.68 \times 10^{-5}$ & $9.43 \times 10^{-5}$ & $9.44 \times 10^{-5}$ \\
\textit{(SE); p} & \scriptsize(3.35e-5); 0.006 & \scriptsize(1.75e-5); $<0.001$ & \scriptsize(1.63e-5); $<0.001$ & \scriptsize(1.52e-5); $<0.001$ & \scriptsize(8.89e-6); $<0.001$ & \scriptsize(8.92e-6); $<0.001$ \\
\midrule
Gender & $1.07 \times 10^{-3}$ & $-5.83 \times 10^{-5}$ & $1.00 \times 10^{-3}$ & $8.54 \times 10^{-4}$ & $6.44 \times 10^{-4}$ & $6.38 \times 10^{-4}$ \\
\textit{(SE); p} & \scriptsize(1.08e-3); 0.322 & \scriptsize(5.25e-4); 0.911 & \scriptsize(4.83e-4); 0.038 & \scriptsize(4.54e-4); 0.060 & \scriptsize(2.70e-4); 0.017 & \scriptsize(2.70e-4); 0.018 \\
\midrule
\textbf{Triage Class} & & & & & & \\
P1 & $3.35 \times 10^{-3}$ & $5.79 \times 10^{-3}$ & $5.45 \times 10^{-3}$ & $5.85 \times 10^{-3}$ & $5.49 \times 10^{-3}$ & $5.64 \times 10^{-3}$ \\
\textit{(SE); p} & \scriptsize(2.76e-3); 0.225 & \scriptsize(9.16e-4); $<0.001$ & \scriptsize(9.16e-4); $<0.001$ & \scriptsize(7.96e-4); $<0.001$ & \scriptsize(5.01e-4); $<0.001$ & \scriptsize(5.00e-4); $<0.001$ \\
P2 & $1.93 \times 10^{-3}$ & $2.30 \times 10^{-4}$ & $-2.28 \times 10^{-4}$ & $3.68 \times 10^{-4}$ & $4.80 \times 10^{-5}$ & $1.96 \times 10^{-4}$ \\
\textit{(SE); p} & \scriptsize(1.15e-3); 0.094 & \scriptsize(3.82e-4); 0.547 & \scriptsize(3.99e-4); 0.569 & \scriptsize(3.10e-4); 0.235 & \scriptsize(2.21e-4); 0.828 & \scriptsize(2.23e-4); 0.379 \\
\midrule
Diastolic BP & $-4.01 \times 10^{-5}$ & $-4.44 \times 10^{-5}$ & $-4.59 \times 10^{-5}$ & $1.06 \times 10^{-5}$ & $-2.21 \times 10^{-5}$ & $-2.29 \times 10^{-5}$ \\
\textit{(SE); p} & \scriptsize(5.92e-5); 0.498 & \scriptsize(2.08e-5); 0.031 & \scriptsize(2.32e-5); 0.048 & \scriptsize(2.17e-5); 0.626 & \scriptsize(1.23e-5); 0.073 & \scriptsize(1.24e-5); 0.064 \\
\midrule
Systolic BP & $3.46 \times 10^{-5}$ & $-1.94 \times 10^{-5}$ & $-4.60 \times 10^{-5}$ & $-7.78 \times 10^{-5}$ & $-4.51 \times 10^{-5}$ & $-4.54 \times 10^{-5}$ \\
\textit{(SE); p} & \scriptsize(4.73e-5); 0.465 & \scriptsize(1.62e-5); 0.230 & \scriptsize(1.47e-5); 0.002 & \scriptsize(1.44e-5); $<0.001$ & \scriptsize(8.64e-6); $<0.001$ & \scriptsize(8.65e-6); $<0.001$ \\
\midrule
Pulse & $2.02 \times 10^{-5}$ & $6.27 \times 10^{-5}$ & $4.97 \times 10^{-3}$ & $6.27 \times 10^{-5}$ & $5.73 \times 10^{-5}$ & $5.62 \times 10^{-5}$ \\
\textit{(SE); p} & \scriptsize(3.04e-5); 0.506 & \scriptsize(1.74e-5); $<0.001$ & \scriptsize(1.59e-5); 0.002 & \scriptsize(1.57e-5); $<0.001$ & \scriptsize(9.00e-6); $<0.001$ & \scriptsize(9.04e-6); $<0.001$ \\
\midrule
Resp.\ Rate & $-1.82 \times 10^{-4}$ & $3.56 \times 10^{-4}$ & $4.12 \times 10^{-4}$ & $7.12 \times 10^{-4}$ & $4.30 \times 10^{-4}$ & $4.17 \times 10^{-4}$ \\
\textit{(SE); p} & \scriptsize(2.60e-4); 0.485 & \scriptsize(3.53e-4); 0.314 & \scriptsize(2.83e-4); 0.146 & \scriptsize(3.24e-4); 0.028 & \scriptsize(1.66e-4); 0.010 & \scriptsize(1.67e-4); 0.012 \\
\midrule
SAO2 & $-1.79 \times 10^{-4}$ & $-2.60 \times 10^{-4}$ & $-2.55 \times 10^{-4}$ & $-3.44 \times 10^{-4}$ & $-2.81 \times 10^{-4}$ & $-2.82 \times 10^{-4}$ \\
\textit{(SE); p} & \scriptsize(1.75e-4); 0.305 & \scriptsize(1.32e-4); 0.048 & \scriptsize(1.48e-4); 0.086 & \scriptsize(1.43e-4); 0.016 & \scriptsize(7.97e-5); $<0.001$ & \scriptsize(7.97e-5); $<0.001$ \\
\botrule
\end{tabular}
\end{table*}

\begin{table*}[!htbp]
\caption[]{Table~\ref{table:RD-realdat} (continued).}
\centering
\footnotesize
\begin{tabular}{lcccccc}
\toprule
\textbf{Variables} & \textbf{MIMIC} & \textbf{SGH-I} & \textbf{SGH-II} & \textbf{SGH-III} & \textbf{Pooled} & \textbf{DiSAH} \\
\midrule
MI & $-4.56 \times 10^{-3}$ & $5.89 \times 10^{-3}$ & $4.92 \times 10^{-3}$ & $5.19 \times 10^{-3}$ & $5.30 \times 10^{-3}$ & $5.31 \times 10^{-3}$ \\
\textit{(SE); p} & \scriptsize(3.60e-3); 0.206 & \scriptsize(1.46e-3); $<0.001$ & \scriptsize(1.45e-3); $<0.001$ & \scriptsize(1.29e-3); $<0.001$ & \scriptsize(8.00e-4); $<0.001$ & \scriptsize(7.99e-4); $<0.001$ \\
\midrule
CHF & $9.82 \times 10^{-3}$ & $8.89 \times 10^{-4}$ & $-2.65 \times 10^{-3}$ & $6.38 \times 10^{-4}$ & $-2.24 \times 10^{-4}$ & $-2.26 \times 10^{-4}$ \\
\textit{(SE); p} & \scriptsize(5.49e-3); 0.858 & \scriptsize(1.10e-3); 0.420 & \scriptsize(9.41e-4); 0.005 & \scriptsize(1.02e-3); 0.531 & \scriptsize(5.88e-4); 0.703 & \scriptsize(5.88e-4); 0.701 \\
\midrule
PVD & $-6.86 \times 10^{-4}$ & $-1.14 \times 10^{-3}$ & $-3.64 \times 10^{-4}$ & $1.66 \times 10^{-4}$ & $-5.17 \times 10^{-4}$ & $-5.15 \times 10^{-4}$ \\
\textit{(SE); p} & \scriptsize(2.64e-3); 0.009 & \scriptsize(1.04e-3); 0.275 & \scriptsize(1.06e-3); 0.730 & \scriptsize(1.01e-3); 0.870 & \scriptsize(5.93e-4); 0.383 & \scriptsize(5.93e-4); 0.385 \\
\midrule
Stroke & $-5.00 \times 10^{-3}$ & $1.07 \times 10^{-3}$ & $5.40 \times 10^{-4}$ & $1.68 \times 10^{-4}$ & $4.58 \times 10^{-4}$ & $5.08 \times 10^{-4}$ \\
\textit{(SE); p} & \scriptsize(2.52e-3); 0.047 & \scriptsize(8.25e-4); 0.193 & \scriptsize(7.31e-4); 0.460 & \scriptsize(6.95e-4); 0.809 & \scriptsize(4.30e-4); 0.286 & \scriptsize(4.29e-4); 0.237 \\
\midrule
Pulmonary & $3.18 \times 10^{-3}$ & $-1.44 \times 10^{-3}$ & $-7.18 \times 10^{-4}$ & $-1.55 \times 10^{-3}$ & $-1.08 \times 10^{-3}$ & $-1.06 \times 10^{-3}$ \\
\textit{(SE); p} & \scriptsize(4.26e-3); 0.456 & \scriptsize(1.04e-3); 0.166 & \scriptsize(1.16e-3); 0.535 & \scriptsize(1.04e-3); 0.135 & \scriptsize(6.13e-4); 0.078 & \scriptsize(6.13e-4); 0.084 \\
\midrule
Renal & $5.45 \times 10^{-4}$ & $5.37 \times 10^{-4}$ & $4.80 \times 10^{-4}$ & $7.78 \times 10^{-4}$ & $5.40 \times 10^{-4}$ & $5.74 \times 10^{-4}$ \\
\textit{(SE); p} & \scriptsize(4.22e-3); 0.897 & \scriptsize(6.72e-4); 0.424 & \scriptsize(6.95e-4); 0.489 & \scriptsize(6.16e-4); 0.206 & \scriptsize(3.78e-4); 0.153 & \scriptsize(3.78e-4); 0.129 \\
\midrule
\textbf{Malignancy} & & & & & & \\
Tumor/Leuk & $2.27 \times 10^{-3}$ & $1.79 \times 10^{-3}$ & $2.81 \times 10^{-3}$ & $1.32 \times 10^{-3}$ & $2.04 \times 10^{-3}$ & $2.07 \times 10^{-3}$ \\
\textit{(SE); p} & \scriptsize(3.39e-3); 0.502 & \scriptsize(9.38e-4); 0.057 & \scriptsize(1.01e-3); 0.005 & \scriptsize(8.18e-4); 0.107 & \scriptsize(5.26e-4); $<0.001$ & \scriptsize(5.26e-4); $<0.001$ \\
Metastatic & $1.35 \times 10^{-2}$ & $6.89 \times 10^{-3}$ & $4.93 \times 10^{-3}$ & $7.84 \times 10^{-3}$ & $6.70 \times 10^{-3}$ & $6.78 \times 10^{-3}$ \\
\textit{(SE); p} & \scriptsize(8.78e-3); 0.124 & \scriptsize(1.16e-3); $<0.001$ & \scriptsize(1.06e-3); $<0.001$ & \scriptsize(1.11e-3); $<0.001$ & \scriptsize(6.45e-4); $<0.001$ & \scriptsize(6.46e-4); $<0.001$ \\
\midrule
\textbf{Liver disease} & & & & & & \\
Mild & $4.60 \times 10^{-3}$ & $-3.82 \times 10^{-4}$ & $1.22 \times 10^{-4}$ & $7.39 \times 10^{-4}$ & $2.72 \times 10^{-4}$ & $2.96 \times 10^{-4}$ \\
\textit{(SE); p} & \scriptsize(6.47e-3); 0.477 & \scriptsize(1.08e-3); 0.722 & \scriptsize(1.06e-3); 0.908 & \scriptsize(1.11e-3); 0.505 & \scriptsize(6.37e-4); 0.670 & \scriptsize(6.36e-4); 0.642 \\
Severe & $5.56 \times 10^{-3}$ & $2.16 \times 10^{-3}$ & $9.25 \times 10^{-3}$ & $6.85 \times 10^{-3}$ & $6.00 \times 10^{-3}$ & $6.00 \times 10^{-3}$ \\
\textit{(SE); p} & \scriptsize(1.29e-2); 0.665 & \scriptsize(2.20e-3); 0.326 & \scriptsize(3.37e-3); 0.006 & \scriptsize(2.36e-3); 0.004 & \scriptsize(1.49e-3); $<0.001$ & \scriptsize(1.49e-3); $<0.001$ \\
\midrule
\textbf{Diabetes} & & & & & & \\
w/o compl.\ & $-2.07 \times 10^{-3}$ & $-8.74 \times 10^{-4}$ & $-2.32 \times 10^{-3}$ & $1.26 \times 10^{-3}$ & $-5.58 \times 10^{-4}$ & $-5.92 \times 10^{-4}$ \\
\textit{(SE); p} & \scriptsize(3.72e-3); 0.577 & \scriptsize(1.49e-3); 0.559 & \scriptsize(1.04e-3); 0.026 & \scriptsize(1.52e-3); 0.408 & \scriptsize(7.94e-4); 0.482 & \scriptsize(7.93e-4); 0.455 \\
w/ compl.\ & $9.01 \times 10^{-3}$ & $-1.65 \times 10^{-3}$ & $-1.56 \times 10^{-3}$ & $-1.79 \times 10^{-3}$ & $-1.47 \times 10^{-3}$ & $-1.39 \times 10^{-3}$ \\
\textit{(SE); p} & \scriptsize(8.44e-3); 0.285 & \scriptsize(6.56e-4); 0.012 & \scriptsize(6.30e-4); 0.013 & \scriptsize(5.41e-4); 0.029 & \scriptsize(3.44e-4); $<0.001$ & \scriptsize(3.47e-4); $<0.001$ \\
\botrule
\end{tabular}
\end{table*}
\end{landscape}

\begin{table*}[!htbp]
\caption{Discrimination performance across methods and sites. C-index, integrated AUC (iAUC), and time-dependent AUC at $t = 7, 14, 21, 28$ days are reported with 95\% confidence intervals for Local, Meta, DiSAH, and Pooled methods.}
\label{table:S4-complete-AUCt}
\setlength{\tabcolsep}{0.1cm}
\begin{center}
\footnotesize
\begin{tabular}{llcccc}
\toprule
\textbf{Site} & \textbf{Metrics} & \textbf{Local} & \textbf{Meta} & \textbf{DiSAH} & \textbf{Pooled} \\ 
\midrule
\multirow{6}{*}{Site 1 (MIMIC)} & C-index & 0.720 (0.281--0.901) & 0.821 (0.645--0.921) & \textbf{0.850} (0.662--0.926) & 0.846 (0.659--0.924) \\
 & iAUC & 0.642 (0.317--0.893) & \textbf{0.726} (0.594--0.862) & 0.706 (0.549--0.851) & 0.704 (0.548--0.850) \\
 & AUC ($t=7$) & 0.608 (0.000--0.926) & 0.772 (0.291--0.921) & \textbf{0.833} (0.301--0.948) & 0.825 (0.301--0.946) \\
 & AUC ($t=14$) & 0.647 (0.055--0.917) & 0.730 (0.588--0.901) & \textbf{0.770} (0.615--0.938) & 0.768 (0.614--0.927) \\
 & AUC ($t=21$) & \textbf{0.727} (0.566--0.902) & 0.643 (0.448--0.897) & 0.621 (0.424--0.878) & 0.619 (0.423--0.866) \\
 & AUC ($t=28$) & 0.566 (0.304--0.793) & \textbf{0.695} (0.460--0.895) & 0.669 (0.416--0.879) & 0.668 (0.416--0.879) \\
\midrule
\multirow{6}{*}{Site 2 (SGH-I)} & C-index & 0.831 (0.791--0.870) & 0.822 (0.775--0.864) & \textbf{0.839} (0.800--0.876) & \textbf{0.839} (0.800--0.876) \\
 & iAUC & 0.754 (0.701--0.804) & 0.733 (0.672--0.789) & \textbf{0.764} (0.708--0.811) & \textbf{0.764} (0.708--0.810) \\
 & AUC ($t=7$) & 0.783 (0.705--0.866) & \textbf{0.804} (0.724--0.875) & 0.802 (0.733--0.877) & 0.801 (0.731--0.876) \\
 & AUC ($t=14$) & 0.739 (0.657--0.805) & 0.717 (0.609--0.810) & \textbf{0.744} (0.643--0.820) & \textbf{0.744} (0.642--0.820) \\
 & AUC ($t=21$) & 0.698 (0.626--0.786) & 0.670 (0.592--0.764) & \textbf{0.713} (0.640--0.794) & \textbf{0.713} (0.638--0.793) \\
 & AUC ($t=28$) & 0.713 (0.577--0.807) & 0.686 (0.522--0.801) & 0.725 (0.591--0.812) & \textbf{0.726} (0.591--0.811) \\
\midrule
\multirow{6}{*}{Site 3 (SGH-II)} & C-index & 0.812 (0.746--0.852) & 0.809 (0.768--0.837) & 0.821 (0.768--0.860) & \textbf{0.839} (0.800--0.876) \\
 & iAUC & 0.728 (0.666--0.779) & 0.731 (0.668--0.782) & 0.741(0.674--0.789) & \textbf{0.764} (0.708--0.810) \\
 & AUC ($t=7$) & 0.737 (0.635--0.867) & 0.734 (0.628--0.843) & 0.749 (0.664--0.863) & \textbf{0.801} (0.731--0.876) \\
 & AUC ($t=14$) & 0.668 (0.550--0.766) & 0.702 (0.604--0.790) & 0.704 (0.599--0.784) & \textbf{0.744} (0.642--0.820) \\
 & AUC ($t=21$) & 0.738 (0.648--0.819) & \textbf{0.748} (0.643--0.832) & 0.744 (0.629--0.824) & 0.713 (0.638--0.793) \\
 & AUC ($t=28$) & 0.756 (0.655--0.837) & 0.758 (0.654--0.833) & \textbf{0.761} (0.666--0.842) & 0.726 (0.591--0.811) \\
\midrule
\multirow{6}{*}{Site 4 (SGH-III)} & C-index & 0.847 (0.816--0.871) & 0.834 (0.800--0.862) & \textbf{0.853} (0.825--0.876) & \textbf{0.853} (0.825--0.876) \\
 & iAUC & 0.789 (0.745--0.823) & 0.782 (0.744--0.819) & \textbf{0.793} (0.751--0.829) & 0.792 (0.751--0.828) \\
 & AUC ($t=7$) & 0.797 (0.752--0.848) & 0.783 (0.732--0.833) & \textbf{0.803} (0.758--0.857) & \textbf{0.803} (0.756--0.857) \\
 & AUC ($t=14$) & 0.697 (0.629--0.766) & \textbf{0.724} (0.661--0.788) & 0.710 (0.638--0.788) & 0.707 (0.635--0.786) \\
 & AUC ($t=21$) & 0.729 (0.662--0.783) & \textbf{0.739} (0.664--0.796) & 0.734 (0.670--0.795) & 0.731 (0.667--0.792) \\
 & AUC ($t=28$) & \textbf{0.728} (0.656--0.778) & \textbf{0.728} (0.639--0.791) & 0.708 (0.628--0.763) & 0.707 (0.625--0.761) \\
\botrule
\end{tabular}
\end{center}
\end{table*}

\end{appendices}

\end{document}